\makeatletter
\newcommand\blfootnote[1]{%
  \begingroup
  \renewcommand\thefootnote{}\footnotetext{#1}%
  \addtocounter{footnote}{1}%
  \endgroup
}
\makeatother 
\documentclass[10pt,twocolumn,letterpaper]{article}

\usepackage{cvpr}              

\usepackage{lineno}

\renewcommand{\paragraph}[1]{\vspace{.5em}\noindent\textbf{#1.}}

\definecolor{cvprblue}{rgb}{0.21,0.49,0.74}
\usepackage[pagebackref,breaklinks,colorlinks,allcolors=cvprblue]{hyperref}
\usepackage{multirow}
\usepackage{booktabs}
\usepackage{array}
\usepackage{graphicx}
\usepackage{tikz}
\usepackage{lineno}
\usepackage{standalone} 
\usepackage[accsupp]{axessibility}  

\def\paperID{ CV4Clinical 37} 
\def\confName{CVPR}
\def\confYear{2026}

\title{Beyond Fluency:  A Clinical Benchmark and Anomaly-Enhanced Baseline for Spine MRI Report Generation}

\author{
Bruno Palau$^{1}$ \and
Franziska Vogt$^{1}$ \and
Daria Laslo$^{1}$ \and
Haobo Li$^{1}$ \and
Ender Konukoglu$^{1}$ \and
\begin{tabular}[t]{@{}c@{\hspace{1.5em}}c@{}}
Maria Monzon$^{1,2\dagger}$ & Catherine R. Jutzeler$^{1,2\dagger}$ \\
\multicolumn{2}{c}{\footnotesize $^\dagger$Shared last authorship}
\end{tabular}
\\
$^{1}$ ETH Zurich, Switzerland\\
$^{2}$Swiss Institute of Bioinformatics (SIB),  Switzerland\\
{\tt\small maria.monzonronda@hest.ethz.ch}
}

\begin{document}
\maketitle

\blfootnote{© 2026 IEEE. Personal use of this material is permitted.
Permission from IEEE must be obtained for all other uses, in any current
for future media, including reprinting or republishing for advertising or promotion, creating collective works, resale or redistribution to servers or lists, or reusing any copyrighted component in other works.}

\begin{abstract}


Radiology reporting is time-consuming and subject to inter-rater variability, making automated report generation an attractive clinical application for Vision-Language Models (VLMs). 
We benchmark state-of-the-art VLMs on lumbar 
spine MRI with a focus on diagnostic accuracy and demonstrate that standard lexical and semantic metrics poorly reflect
 clinical correctness: fluent, well-structured reports can score 
highly while containing clinically meaningful diagnostic errors.
To address this failure mode, we propose an architecture-agnostic framework that augments VLM inputs with spatially localized, disc-level anomaly heatmaps generated by a semi-supervised U-Net++ model. 
These heatmaps both improve anatomical sensitivity through explicit visual grounding and provide an independent interpretability output for clinical oversight, moving us closer to diagnostically reliable, visually grounded VLMs for lumbar spine MRI interpretation.

\end{abstract}

\section{Introduction}
 \label{sec:intro}

Low back pain (LBP) affects approximately 619 million individuals worldwide and is a leading cause of years lived with disability across all age groups~\cite{hartvigsen_what_2018}. 
Lumbar spine MRI is the gold standard for non-invasive assessment of LBP, providing exceptional soft tissue contrast to evaluate inter-vertebral discs, neural foramina, and degenerative changes \cite{saeed_research_2024}. 
In routine clinical practice, radiological findings are communicated through radiology reports that summarize the presence, severity, and anatomical distribution of pathological abnormalities.
However, manual radiology reporting is time-consuming and subject to inter-rater variability~\cite{reiner_challenges_2009, brady_error_2017}.

\begin{figure}[!t]
  \centering
   \includegraphics[width=0.98\linewidth]{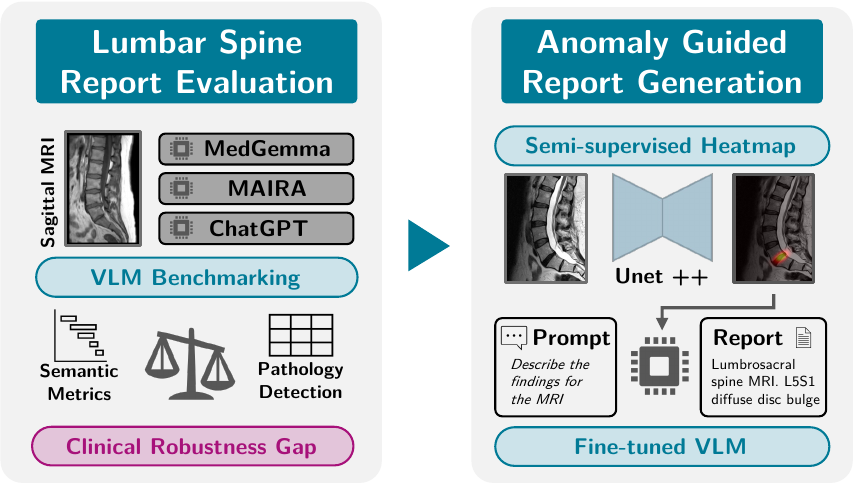}
   \caption{Graphical overview of the benchmarking and
anomaly-guided report generation framework}
   \label{fig:main-graphical-overview}
\end{figure} 
Automated radiology report generation has emerged as a promising solution, leveraging deep learning to translate medical images into diagnostic text. 
Despite rapid progress~\cite{ryu_vision-language_2025, casey_systematic_2021}, existing benchmarks focus  primarily on chest X-ray and CT imaging~\cite{dishner_survey_2024}, with limited attention to MRI. Even when MRI is included 
in automated reporting systems, efforts have primarily targeted brain 
imaging~\cite{dishner_survey_2024}, leaving musculoskeletal applications—particularly 
lumbar spine MRI—comparatively underexplored~\cite{dishner_survey_2024, 
ryu_vision-language_2025}.

Addressing this gap is clinically critical yet technically challenging. 
Unlike the relative homogeneity of radiographs, lumbar spine MRI 
introduces substantial acquisition- and scanner-dependent variability, 
with significant differences in field of view, sequence type (T1w/T2w), 
slice thickness, and intensity distributions~\cite{dishner_survey_2024, 
bluemke_assessing_2020}.
Furthermore, lumbar spine reporting requires precise  localization of degenerative pathologies across six distinct 
intervertebral levels (T12-L1 through L5-S1),  presenting a unique structural 
challenge for report generation models.

Beyond domain-specific imaging challenges, a critical bottleneck lies in model evaluation. Current benchmarks rely predominantly on generic  language metrics (e.g., BLEU, BERTScore) that measure lexical and 
semantic similarity to reference text. While effective at assessing  linguistic 
fluency, these metrics are fundamentally misaligned with the diagnostic reality  of lower back pain (LBP) assessment. 
Clinical guidelines prioritise 
the rapid triage of structural pathologies (e.g., disc herniation,  stenosis) and rely strictly on exact, disc-level spatial  localization~\cite{Bardin2017-LBP}. Consequently, a model that  fluently describes a finding but assigns it to the wrong vertebral level commits a severe clinical failure. However, standard text metrics 
routinely fail to penalise this spatial mismatch. 
This evaluation gap highlights a potential limitation for clinical deployment ~\cite{topol_high-performance_2019, casey_systematic_2021} and motivates benchmarks that explicitly assess diagnostic localization and image grounding.

In this work, we conduct an  evaluation of state-of-the-art vision-language models (VLM) for lumbar spine MRI report generation, addressing three fundamental gaps: the absence of  rigorous benchmarks for foundation VLMs on spine MRI, the failure of standard evaluation metrics to capture clinical correctness in LBP assessment, and the poor visual grounding of foundation models outside their primary training domains. Our contributions are:

\textbf{(i) Benchmarking VLMs on lumbar spine MRI with limited data.} We evaluate general-purpose and medically specialized multimodal models~\cite{openai_chatgpt_2025, bannur_maira-2_2024, sellergren_medgemma_2025, nath_vila-m3_2025} using complementary protocols: (a) free-text report generation measured against clinical reference reports, and (b) structured per-finding and per-level diagnostic classification measured against radiological gradings. 
We further evaluate fine-tuning on a limited spine MRI dataset, examining whether metric-driven optimization improves diagnostic accuracy.  

\textbf{(ii) Robustness analysis of models and metrics under clinically representative variability.} We assess model sensitivity to MRI contrast (T1w, T2w), central slice positioning, resampling, and prompt formulation to determine whether apparent performance stability reflects robustness. 
We also conduct a controlled perturbation analysis, introducing diagnostic negations, anatomical swaps, terminology variations, and structural changes to assess whether standard language-generation metrics adequately penalize critical clinical errors.

\textbf{(iii) Anomaly-guided report generation via explicit visual grounding.} 
We propose a disc-level anomaly detection framework that produces spatially localized heatmaps of deviations from normal anatomy. 
By integrating these heatmaps as an 
auxiliary visual input, we provide the VLM with explicit, image-derived 
spatial evidence. This architecture-agnostic approach serves a dual 
purpose: it enhances diagnostic reasoning by directing model attention to 
pathological regions, and it provides an inspectable visual trace to 
support clinical interpretability.


\section{Related Work}
\paragraph{Radiology Report Generation (RRG)}   aims to bridge the gap between visual perception and clinical documentation~\cite{RrgReview,monshi_deep_2020, li_llava-med_2023}, and sometimes considered a specialized case of the image captioning task.
Early systems used retrieval-based matching or rigid templates, providing consistency but lacking the flexibility needed for complex diagnostics.
Modern approaches 
treat RRG as a vision-to-language translation problem. Encoder-decoder 
architectures, convolutional networks~\cite{chen-etal-2021-cross-modal} 
or vision transformers~\cite{nicolson-improving-cvt2distilgpt2, 
wang2023metransformer, liu2024in-context-acmmm}, extract visual features 
and fuse them with contextual inputs such as prior reports or clinical 
indications. Contrastive pretraining~\cite{radford2021l-CLIP} has further 
improved the semantic alignment between radiological images and clinical 
terminology~\cite{wang2022-GIT}, while powerful autoregressive language 
models~\cite{aaai-liu2024bootstrapping-llm, liu2024in-context-acmmm, 
2024-iclr-cxr-llm, wang-2023-r2gengpt} decode these representations into 
coherent free-text reports.

Despite rapid progress~\cite{casey_systematic_2021, wu_vision-language_2025, sloan_automated_2025}, existing benchmarks focus on chest X-ray and CT data~\cite{hamamci_ct2rep_2024, bannur_maira-2_2024}.
Work on spine MRI report generation remains sparse, with very few publicly 
available models~\cite{LewandrowskI2020, Park2026-AISpine}.
Early efforts 
explored weakly supervised frameworks utilizing object-level annotations to 
localize vertebrae prior to text generation~\cite{Han2018-RGS}. Recently fine-tuned GIT~\cite{wang2022-GIT} on composite axial MRI, and LLM-based structured reporting \cite{Park2024-RRG} with label extraction \cite{Park_2024-ASL} have been explored for spine reports.
The limited works underscore the difficulty of obtaining structured annotations in a field dominated by free-text documentation and variable reporting styles~\cite{MartnNoguerol2024}, motivating automated methods that robustly achieve spatial grounding without dense pixel-level supervision.

\paragraph{Vision-Language Foundation Models} 
AI models for RRG are rapidly advancing~\cite{ryu_vision-language_2025}, driven by multimodal VLMs~\cite{ryu_vision-language_2025, luo_vividmed_2025}, which align visual encoders with large language decoders to enable flexible, prompt-based task specification. 
Recent models applied to clinical contexts span a spectrum from broad, domain-agnostic generalists, such as the lightweight BiomedGPT~\cite{zhang_generalist_2024} and the proprietary ChatGPT-5.0~\cite{openai_chatgpt_2025}, to dedicated healthcare foundation models. 
Within the specialized domain, models like MedGemma~\cite{sellergren_medgemma_2025} and the agentic VILA-M3~\cite{nath_vila-m3_2025} leverage domain-specific pretraining and medical image encoders to support diverse clinical multi-tasking. 
Furthermore, architectures like MAIRA-2~\cite{bannur_maira-2_2024} have begun incorporating structured input interfaces and spatial grounding via bounding boxes, though they remain primarily unoptimized for MRI.
Indeed, current VLMs often lack reliable inter-vertebral grounding on lumbar spine MRI without dense supervision (e.g., bounding boxes), motivating semi-supervised methods that inject explicit spatial priors into the visual context for clinically reliable reporting~\cite{bannur_maira-2_2024, Han2018-RGS}.

\section{Methods}
\begin{figure*}
  \centering
  \includegraphics[width=0.95\linewidth]{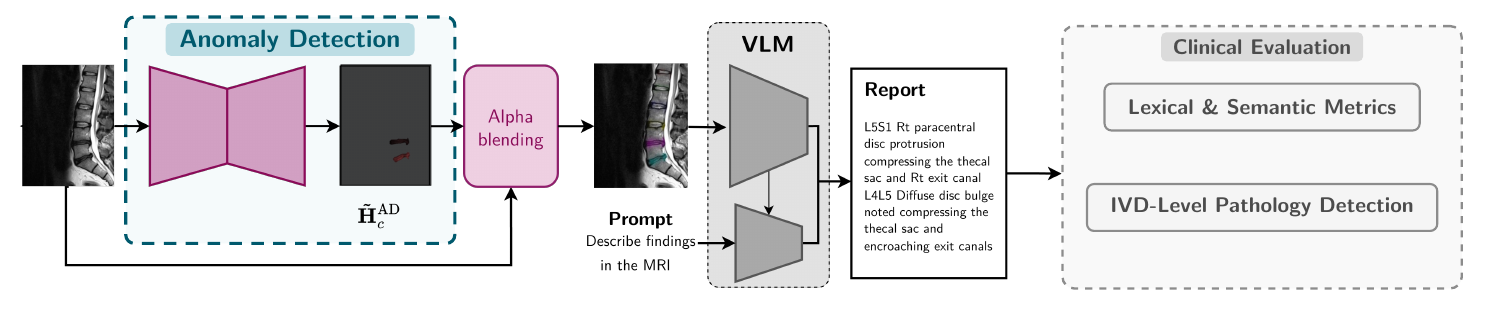}

    \includegraphics[width=0.98\linewidth]{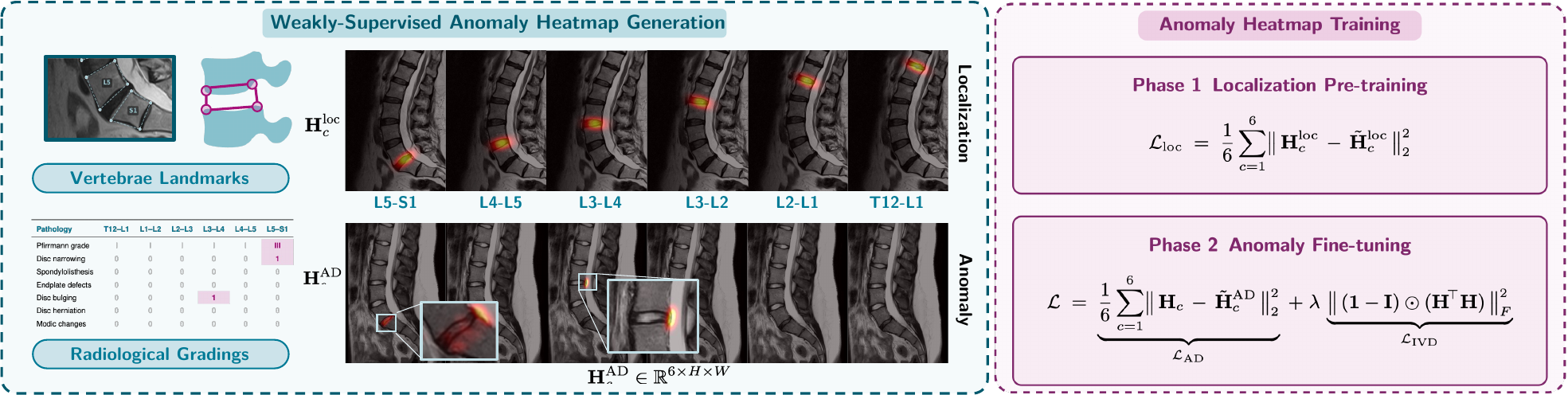}

\caption{\textbf{(Top) Benchmarking and anomaly-guided report generation  framework.} State-of-the-art VLMs are evaluated on free-text report generation  and structured per-finding IVD-level classification. A semi-supervised U-Net++ 
anomaly detector generates disc-level heatmaps that serve dual purpose: 
augmenting VLM inputs with spatially localized image-derived evidence, and 
providing an independent interpretability output for clinical oversight. \newline
\textbf{(Bottom) Heatmap generation and VLM integration.} Per-class heatmap 
channels ($H_1,\dots,H_6$) are collapsed into a single RGB overlay by 
assigning each pixel the color of the maximally activated disc level 
($\arg\max_{c} H_c$) scaled by its activation magnitude ($\max_{c} H_c$), 
then alpha-blended with the original MRI slice and passed as auxiliary visual 
input alongside the original image.}

  \label{fig:main-methods-overview}
\end{figure*}

\subsection{Foundation VLMs for Spine Report Generation}

\paragraph{Model Selection}
We evaluate five multimodal architectures selected to span the design space 
relevant to clinical report generation: a general-purpose lightweight model 
(\textit{BiomedGPT}~\cite{zhang_generalist_2024}), a large-scale proprietary 
generalist (\textit{ChatGPT-5.0}~\cite{openai_chatgpt_2025}), a domain-specific 
model with structured input and spatial grounding 
(\textit{MAIRA-2}~\cite{bannur_maira-2_2024}), a medical foundation model with 
dedicated image encoder (\textit{MedGemma}~\cite{sellergren_medgemma_2025}), 
and an agentic multi-task medical VLM 
(\textit{VILA-M3}~\cite{nath_vila-m3_2025}). 
All models receive a central sagittal MRI slice, considered sufficient for pathology identification and a task prompt requesting a radiological report.

\paragraph{Input Robustness Evaluation}
To test whether reported performance reflects sensitivity to input variation, we perform targeted ablations on visual and textual inputs at inference.
Starting from a default configuration (single central sagittal T2-weighted slice, medium-length 
text prompt), we systematically vary the following parameters: MRI contrast (T1-weighted; combined 
T1-weighted/T2-weighted provided as separate channels), sagittal slice position relative to the 
central slice (center~$\pm1$), image resolution (native versus resampled to 0.5\,mm 
in-plane voxel spacing), and prompt verbosity.
These variations reflect realistic  acquisition- and 
protocol-level variability encountered in routine clinical practice, helping distinguish true robustness from mere apparent stability due to input insensitivity.

\paragraph{Supervised Fine-Tuning of Best Model}
We fine-tune on a deliberately restricted data regime designed to approximate realistic clinical annotation resource limitations, to investigate whether limited domain-specific supervision can narrow the gap between superficial fluency and clinically grounded diagnosis. 
We adopt QLoRA~\cite{dettmers_qlora_2023} to enable parameter-efficient
adaptation without full-model retraining.
Training relies exclusively on the standard next-token prediction
objective, so anatomical localization is learned implicitly
from token likelihood conditioned on the visual input, without any
task-specific loss.
Supervision matches the two label formats used during zero-shot
benchmarking: (i)~grading-derived report text and
(ii)~structured per-finding level. 


\subsection{Robustness for Clinical Applications.}

Standard report generation metrics assess linguistic fluency but not
 diagnostic correctness, a critical gap for clinical
deployment where fluent yet anatomically incorrect outputs pose 
safety risks. We propose two complementary evaluation strategies that
probe clinical robustness beyond surface-level text similarity.

\paragraph{ Controlled Perturbations of Spine Reports}
To assess whether standard language-generation metrics reflect clinical correctness, we systematically perturb reference reports to mimic variations commonly encountered in clinical practice. The proposed perturbations (detailed in Sup. \ref{app:metric-perturbations}) fall into two categories:

\noindent\emph{(i) Lexical variations preserving clinical meaning:}
\textbf{Paraphrasing} replaces words with synonyms while maintaining
diagnostic content (e.g., ``disc bulge'' $\rightarrow$ ``disc protrusion'');
\textbf{boilerplate insertion} prepends standardized headers
 without altering diagnostic
statements; \textbf{report-length inflation} duplicates sentences,
adding redundancy without introducing new findings.
Perturbations preserve clinical diagnostic meaning  by design, and are thus expected to incur minimal metric degradation.

\noindent\emph{(ii) Semantic variations altering diagnostic meaning:}
\textbf{Negation perturbations} insert or remove negation terms to reverse
diagnostic polarity (e.g., ``disc herniation is present''
$\rightarrow$ ``no disc herniation is present'');
\textbf{anatomical location exchanges} substitute vertebral levels while
preserving sentence structure (e.g., ``L4-L5'' $\rightarrow$ ``L2-L3'');
\textbf{medical terminology exchanges} substitute
pathologically distinct terms
(e.g., ``disc herniation`` $\leftrightarrow$ ``disc bulge``,
``stenosis`` $\leftrightarrow$ ``narrowing``). 
Each perturbation introduces a clinically significant diagnostic error, 
probing whether metrics penalize 
meaning changes in proportion to their clinical severity.


\paragraph{Pathological Indication at IVD Level as a Metric}

To directly assess diagnostic grounding, we derive binary presence/absence
labels for each finding at each IVD level (T12-L1 through L5-S1) from
grading-derived structured reports, and prompt models to output only the
affected levels in a constrained format.
This formulation separates anatomical localization ability from free-text 
generation quality, providing a clinically meaningful complement to standard 
language metrics: a model may score highly on lexical and semantic similarity 
while failing to correctly localize pathology at the disc level, exposing  reliance on linguistic priors rather than image-derived evidence.

\subsection{Anomaly Detection-Guided Report Generation}
Since pixel-level anomaly annotations are not available at scale for 
lumbar spine MRI, we derive weakly supervised spatial targets from 
structured per-IVD grading labels to train the detection module.

\paragraph{Disc-Level Heatmap Formulation}
To provide explicit visual grounding, we develop a disc-level anomaly
detection module that generates a six-channel spatial heatmap tensor
$\mathbf{H} \in \mathbb{R}^{6 \times H \times W}$, where each channel
$\mathbf{H}_c$ corresponds to one IVD level
$c \in \{\text{T12-L1},\, \text{L1-L2},\, \text{L2-L3},\,
\text{L3-L4},\, \text{L4-L5},\, \text{L5-S1}\}$,
highlighting deviations from normal anatomy at that level
(Fig.~\ref{fig:main-methods-overview}).
Using vertebrae detection landmarks from SpineNetV2~\cite{windsor_spinenetv2_2022}
and SPIDER gradings, we derive two complementary supervision targets:
(1)~\emph{localization heatmaps} $\tilde{\mathbf{H}}^{\mathrm{loc}}$
covering the disc space and adjacent endplates, and
(2)~\emph{anomaly heatmaps} $\tilde{\mathbf{H}}^{\mathrm{AD}}$
encoding pathology-specific spatial distributions (e.g., posterior disc 
margin for herniation, central canal for stenosis), formulated as soft
spatial distributions with per-finding and per-channel normalization.

\paragraph{Model and Loss.}
UNet++~\cite{zhou_unet_2018,cardoso_monai_2022} predicts $\mathbf{H}$
from three central sagittal slices. The training objective integrates a channel-balanced mean squared error (MSE) with a channel-wise regularization term ($\mathcal{L}_{IVD}$), which restricts the model to constrain the location of IVD levels.
\begin{equation}
  \mathcal{L} =
    \frac{1}{6}\sum_{c=1}^{6}
    \|\mathbf{H}_c - \tilde{\mathbf{H}}_c\|_2^2
    + \lambda\,\|\mathbf{(1-I)}\odot(\mathbf{H}^\top\mathbf{H})\|_F^2,
  \label{eq:main-loss}
\end{equation}
where $\mathbf{M}=\mathbf{1}-\mathbf{I}$ masks off-diagonal entries
to penalize cross-channel co-activation between IVD levels. 
Training proceeds in two phases, with supervision gradually shifted 
from localization targets $\tilde{\mathbf{H}}^{\mathrm{loc}}$ to 
anomaly targets $\tilde{\mathbf{H}}^{\mathrm{AD}}$ via linear interpolation, where $t$ increases monotonically over training epochs:
\begin{equation}
  \tilde{\mathbf{H}}(t) = (1-t)\,\tilde{\mathbf{H}}^{\mathrm{loc}} 
  + t\,\tilde{\mathbf{H}}^{\mathrm{AD}}, \quad t \in [0,1],
  \label{eq:main-curriculum}
\end{equation}

\paragraph{Heatmap Integration into Report Generation.}
Predicted anomaly heatmaps are converted to RGB overlays by assigning each pixel
the color of its maximally activated channel, scaled by activation
magnitude (Fig.~\ref{fig:main-methods-overview}):
\begin{equation}
  \mathbf{S} = \max_c H_c \cdot \mathrm{Color}(\arg\max_c H_c),
\end{equation}
then alpha-blended with the central MRI slice. 
Models supporting multi-image conditioning (ChatGPT, VILA-M3) receive both the original slice and the overlay as separate visual inputs, with prompts adapted to describe the heatmap encoding. 
For models without multi-image support (MedGemma), a structured 
finding report is extracted from the heatmap via a dedicated prompt 
and passed as textual context, preserving the grounding signal 
within the model's input interface.


\section{Experiments}

\begin{table*}[tbh]
\centering
\resizebox{\linewidth}{!}{%
\setlength{\tabcolsep}{4pt}
\begin{tabular}{lllcccccccc}
\toprule
\multicolumn{3}{c}{} &
\multicolumn{4}{c}{\textbf{LSMRI (Free Text Report)}} &
\multicolumn{4}{c}{\textbf{SPIDER (Structured Grading Report)}} \\
\cmidrule(lr){4-7}\cmidrule(lr){8-11}
Model & Size & Contrast &
BERTScore$\uparrow$ & BLEU-4$\uparrow$ & METEOR$\uparrow$ & ROUGE-L$\uparrow$ &
BERTScore$\uparrow$ & BLEU$\uparrow$ & METEOR$\uparrow$ & ROUGE-L$\uparrow$ \\
\midrule

\multirow{6}{*}{BiomedGPT}
  & \multirow{3}{*}{Base}   & T1w     & 88.80 & 0.09 &  4.83 &  9.93 & 83.12 & 0.00 &  1.16 &  3.42 \\
  &                         & T1w+T2w & 88.53 & 0.04 &  2.89 &  5.75 & 87.48 & 0.00 &  1.35 &  2.79 \\
  &                         & T2w     & 88.50 & 0.10 &  5.48 & 10.43 & 86.64 & 0.00 &  1.46 &  4.32 \\
\cline{2-11}
  & \multirow{3}{*}{XLarge} & T1w     & 22.41 & 0.00 &  1.09 &  2.08 & 33.56 & 0.00 &  0.38 &  1.16 \\
  &                         & T1w+T2w & 85.79 & 0.06 &  2.38 &  4.79 & 81.29 & 0.00 &  0.47 &  1.39 \\
  &                         & T2w     & 47.93 & 0.02 &  2.47 &  4.73 & 51.90 & 0.00 &  0.69 &  2.07 \\
\midrule

\multirow{3}{*}{ChatGPT-5.0}
  & \multirow{3}{*}{--}     & T1w     & 91.14 & 0.51 & 21.43 & 11.74 & 92.15 & 0.88 & 19.08 & 16.64 \\
  &                         & T1w+T2w & 91.05 & 0.36 & 21.54 & 11.72 & 91.93 & 0.62 & 18.90 & 15.91 \\
  &                         & T2w     & 91.11 & 0.50 & 21.28 & 11.58 & 92.21 & 0.89 & 19.81 & 16.91 \\
\midrule

\multirow{3}{*}{MAIRA-2}
  & \multirow{3}{*}{--}     & T1w     & 89.57 & 0.14 &  6.97 &  7.29 & 90.48 & 0.89 &  8.87 &  7.32 \\
  &                         & T1w+T2w & 89.00 & 0.10 &  5.33 &  8.46 & 89.97 & 0.56 &  7.91 &  6.21 \\
  &                         & T2w     & 89.73 & 0.21 &  7.54 &  7.48 & 90.50 & 0.92 &  8.69 &  7.09 \\
\midrule

\multirow{6}{*}{MedGemma}
  & \multirow{3}{*}{4B}     & T1w     & 91.09 & 0.66 & 14.21 & 14.25 & 91.76 & 0.38 &  9.82 &  9.33 \\
  &                         & T1w+T2w & 91.16 & 0.61 & 15.27 & 15.87 & 91.71 & 0.33 &  8.76 &  8.77 \\
  &                         & T2w     & 91.12 & 0.55 & 14.91 & 15.24 & 91.83 & 0.36 &  9.63 &  9.37 \\
\cline{2-11}
  & \multirow{3}{*}{27B}    & T1w     & 91.59 & 1.10 & 22.67 & 16.27 & 92.54 & 2.63 & 18.36 & 15.11 \\
  &                         & T1w+T2w & 91.54 & 0.97 & 22.67 & 16.67 & 91.74 & 1.48 & 13.51 & 11.41 \\
  &                         & T2w     & 91.60 & 1.31 & 22.96 & 16.40 & 92.33 & 2.44 & 17.18 & 14.03 \\
\midrule

\multirow{6}{*}{VILA-M3}
  & \multirow{3}{*}{3B}     & T1w     & 88.95 & 0.04 &  3.79 &  4.48 & 89.89 & 1.58 &  8.02 &  5.02 \\
  &                         & T1w+T2w & 89.10 & 0.07 &  5.94 &  6.33 & 89.93 & 1.29 &  7.80 &  5.48 \\
  &                         & T2w     & 88.90 & 0.03 &  5.11 &  4.84 & 89.94 & 1.68 &  8.41 &  5.28 \\
\cline{2-11}
  & \multirow{3}{*}{13B}    & T1w     & 87.22 & 0.06 &  2.42 &  3.92 & 88.39 & 0.16 &  2.87 &  3.42 \\
  &                         & T1w+T2w & 88.63 & 0.09 &  2.55 &  4.81 & 88.39 & 0.01 &  1.61 &  2.09 \\
  &                         & T2w     & 87.25 & 0.01 &  2.32 &  3.90 & 88.14 & 0.00 &  2.64 &  3.63 \\
\bottomrule
\end{tabular}
}
\caption{Zero-shot report generation performance under different MRI contrast configurations.   All metrics are on 0-100 scale (\%).}
\label{tab:main-report-generation-results}
\end{table*}

\subsection{Experimental Setup}

\paragraph{Datasets}
We use three publicly available lumbar spine MRI datasets differing in
acquisition protocol, annotation type, and diagnostic focus.
\textit{(i)~LSMRI}: 515 patients with symptomatic low back pain paired
with expert-written radiology reports; sagittal and axial T1w/T2w
sequences available for most subjects~\cite{sudirman2019lumbar}.
\textit{(ii)~SPIDER}: 218 patients with sagittal T1w/T2w MRI annotated
with structured IVD-level grading labels covering degenerative findings ~\cite{vandergraaf2023spider};
 reference reports were synthesized from grading
matrices using predefined sentence templates with limited lexical
variability to reduce evaluation bias toward any single linguistic
formulation (Sup. \ref{app:spider-report-construction}). 
\textit{(iii)~LumbarDISC}: 1{,}976 patients with sagittal T1w/T2w and axial T2w MRI, annotated with the location and severity of five conditions related to stenosis in five IVD levels~\cite{richards2025lumbardisc}.

\begin{figure*}[htb]
  \centering
  \includegraphics[width=\linewidth]{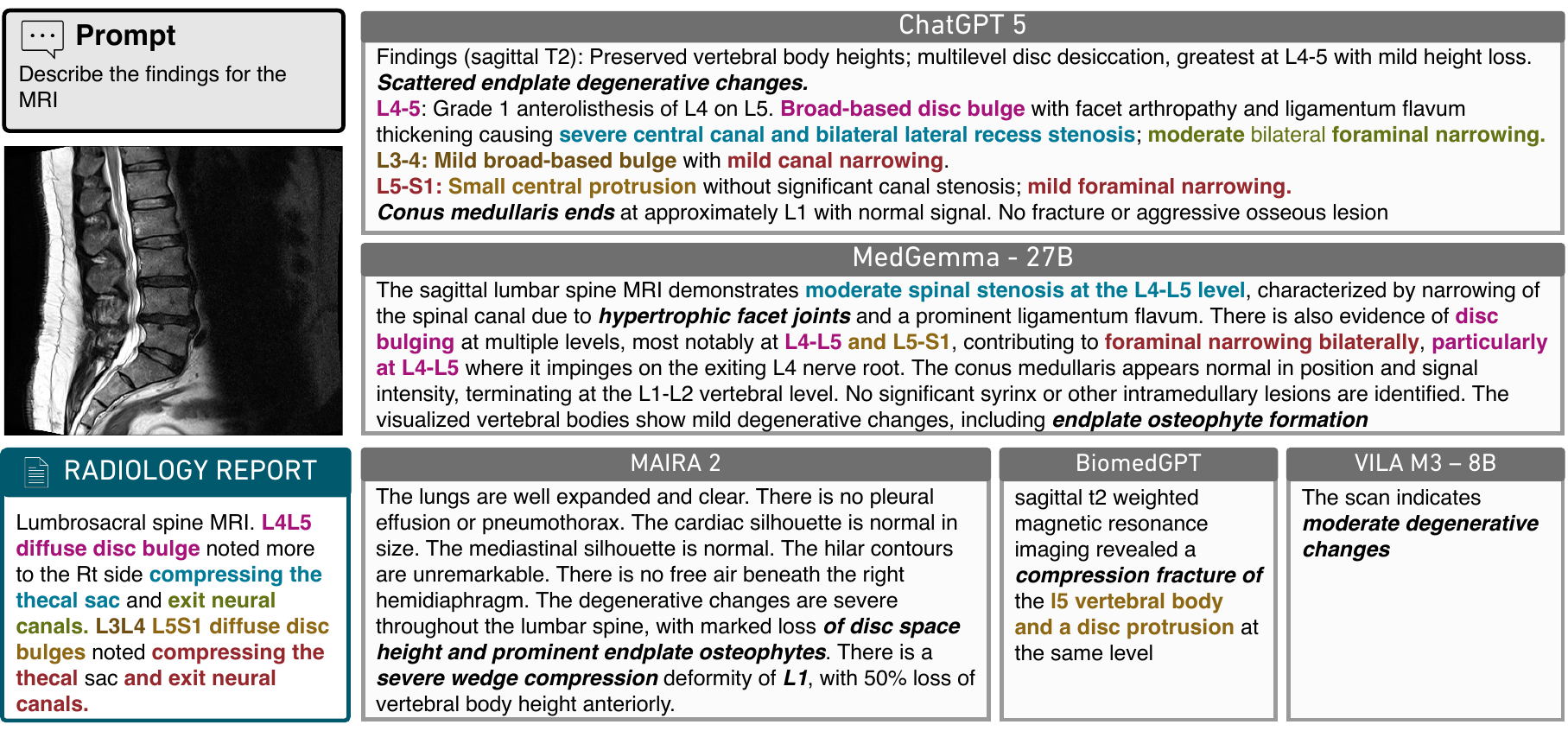}
  \caption{Zero-shot report generation and example outputs with identical inputs. The central sagittal slice and a fixed instruction are provided to each model (BiomedGPT-Base, ChatGPT-5.0, MAIRA-2, MedGemma-27B, VILA-M3-8B), along with the clinical reference report. The outputs show clear differences in clinical specificity, anatomical localization, and format across model families.}
\end{figure*}

\paragraph{Implementation Details}
Zero-shot evaluation covers \textit{BiomedGPT} (Base, Large; 33M--930M 
parameters), \textit{MedGemma} (4B, 27B), \textit{VILA-M3} (3B, 8B, 13B), 
\textit{MAIRA-2}, and \textit{ChatGPT-5.0}. MAIRA-2 receives structured inputs 
via its native interface (indication, technique, and optionally localization  fields).
ChatGPT-5.0 is evaluated in zero-shot only.  
 For models producing free-form outputs (MedGemma, 
ChatGPT-5.0, VILA-M3), generated text was delimited by 
\texttt{[[REPORT\_START]]} and \texttt{[[REPORT\_END]]} tokens and extracted 
deterministically, isolating clinical content from model-specific preambles or 
closing statements.
All volumes were reoriented and 
intensity-normalized; slices were aligned to a consistent spine orientation 
and clipped at the 98th percentile. Unless stated otherwise, models receive a  single central sagittal T2-weighted slice. 

\textit{Robustness experiments.} All ablations additionally evaluate 
T1-weighted and combined T1w/T2w inputs, as well as a three-slice 
stack (center~$\pm 1$), applied to MedGemma and ChatGPT-5.0 on both 
LSMRI and SPIDER.

\textit{VLM fine-tuning.}
MedGemma-4B is fine-tuned with QLoRA~\cite{dettmers_qlora_2023}, keeping
base weights frozen in 4-bit NF4 form and training low-rank adapters in
bfloat16 (20.4M parameters, 0.47\% of total). 
Training uses AdamW with cosine scheduling, linear warmup over 3\% of
steps, and gradient accumulation of 16 steps. Selected hyperparameters:
$\text{lr}=2^{-4}$, $r=32$, $\alpha=64$.
Fine-tuning uses an 80/20 patient-level split on SPIDER with both T1w
and T2w scans; and is evaluated
on the held-out LSMRI test set.

\textit{Anomaly detection training.}
UNet++ is pre-trained for disc localization on 85\% of LumbarDISC and 
30\% of LSMRI for 351 epochs, then fine-tuned on anomaly heatmaps using 
SPIDER for 501 epochs, with $\lambda_t$ ramped linearly from 0 to 1 over 
1{,}000 training steps. Both phases use AdamW with a fixed learning rate 
of $10^{-3}$. Data augmentation includes random affine transformations 
(rotation, translation, scaling), additive Gaussian noise, random zoom, 
and random Gaussian smoothing, applied consistently across both training  phases.

\paragraph{Evaluation Metrics}
\textit{Report generation:} BLEU-4~\cite{papineni_bleu_2001},
ROUGE-L~\cite{lin_rouge_nodate},
METEOR~\cite{noauthor_proceedings_nodate}, and
BERTScore F1~\cite{zhang_bertscore_2020}. 
\textit{Anatomical localization:} sensitivity, specificity, balanced
accuracy, F1, and accuracy, reported per-finding and per-IVD-level to
expose anatomical biases.
\textit{Anomaly detection:} DICE (threshold 0.1)
RMSE on $[0,1]$-normalized heatmaps, stratified by IVD
level and reported as mean $\pm$ SD.

\subsection{Zero-Shot Report Generation}

Table~\ref{tab:main-report-generation-results} summarizes zero-shot performance across 
all models. MedGemma-27B and ChatGPT-5.0 achieve the highest scores on both 
datasets; BiomedGPT and smaller VILA-M3 variants show near-zero BLEU, 
consistent with degenerate outputs. 

\textbf{Input Robustness Ablation.} No input ablation-prompt length, MRI 
contrast, slice selection, or resampling-yielded consistent improvements 
(BERTScore~$\Delta{\leq}\pm0.2$, BLEU~$\Delta{\leq}\pm3.7$), with gains on 
one metric coinciding with losses on another (Sup.~\ref{tab:supp-slice-ablation-results}).

\begin{figure*}[tbh]
  \centering
  \begin{subfigure}{0.42\linewidth}
    \includegraphics[width=\linewidth]{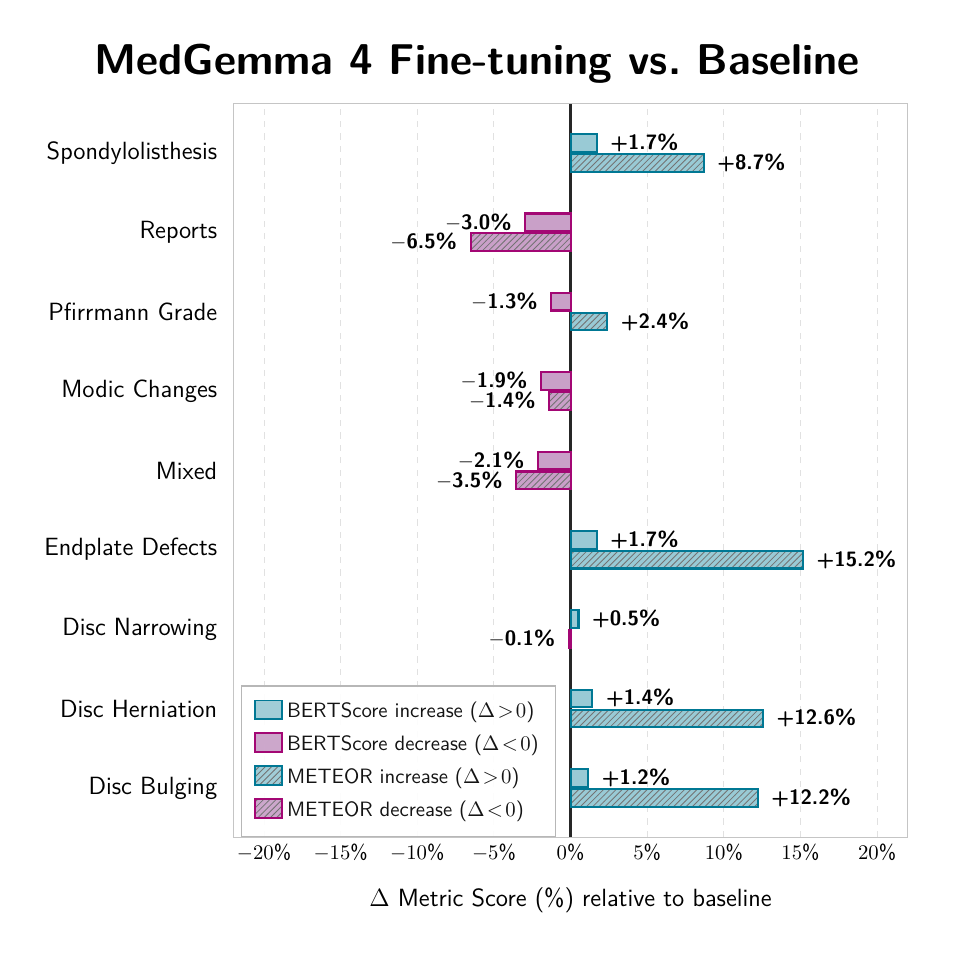}
    \caption{Fine-tuning strategy ablation  study }
    \label{fig:main-finetuning-ablation}

  \end{subfigure}
  \hfill
  \begin{subfigure}{0.4\linewidth}
    \includegraphics[width=\linewidth]{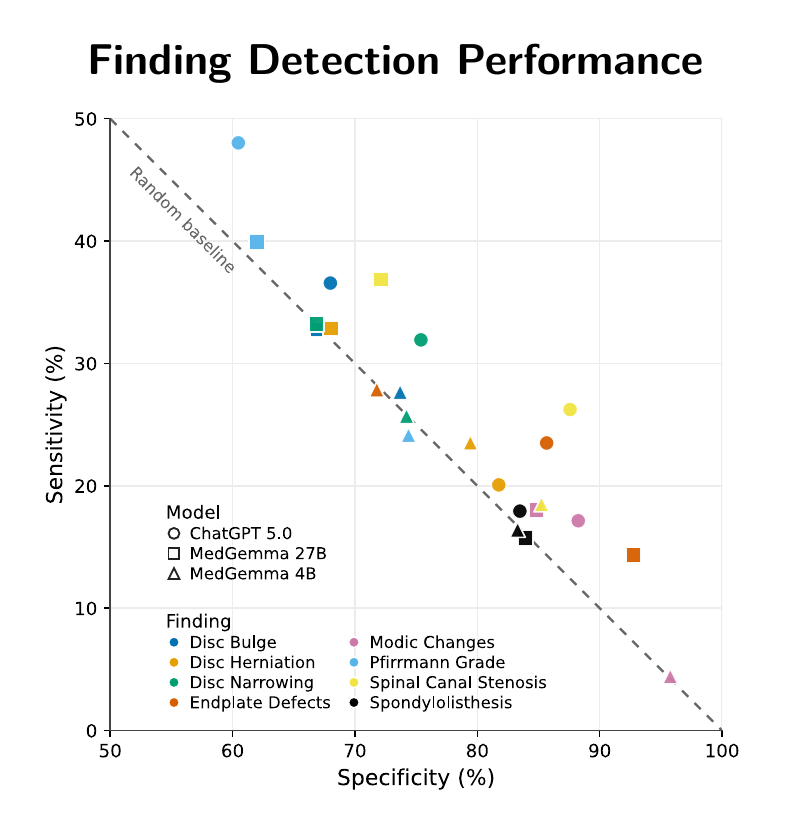}
    \caption{Per-finding classification performance.}

    \label{fig:main-classification-findings}

  \end{subfigure}
  \caption{Level-wise classification on SPIDER/LumbarDISC shown as
  sensitivity--specificity trade-offs aggregated 
  per finding (right). Marker shapes denote models; colors denote
  classes; the dashed diagonal indicates random performance.}
\end{figure*}

\textbf{Supervised Fine-Tuning.}
Fine-tuning MedGemma-4B with QLoRA on finding-based supervision (per-IVD tokens) consistently improved metrics over the pre-trained baseline across all findings (Fig.~\ref{fig:main-finetuning-ablation}a), 
despite no explicit report-generation objective. C
onversely, fine-tuning 
directly on report text degraded performance and increased output 
variability, including anatomically implausible references to 
non-existent vertebral levels (e.g., L6--L7), despite such labels 
never appearing during training. 

\subsection{Metric Robustness Analysis}
\begin{figure}[tbh]
    \centering
    \includegraphics[width=0.8\linewidth]{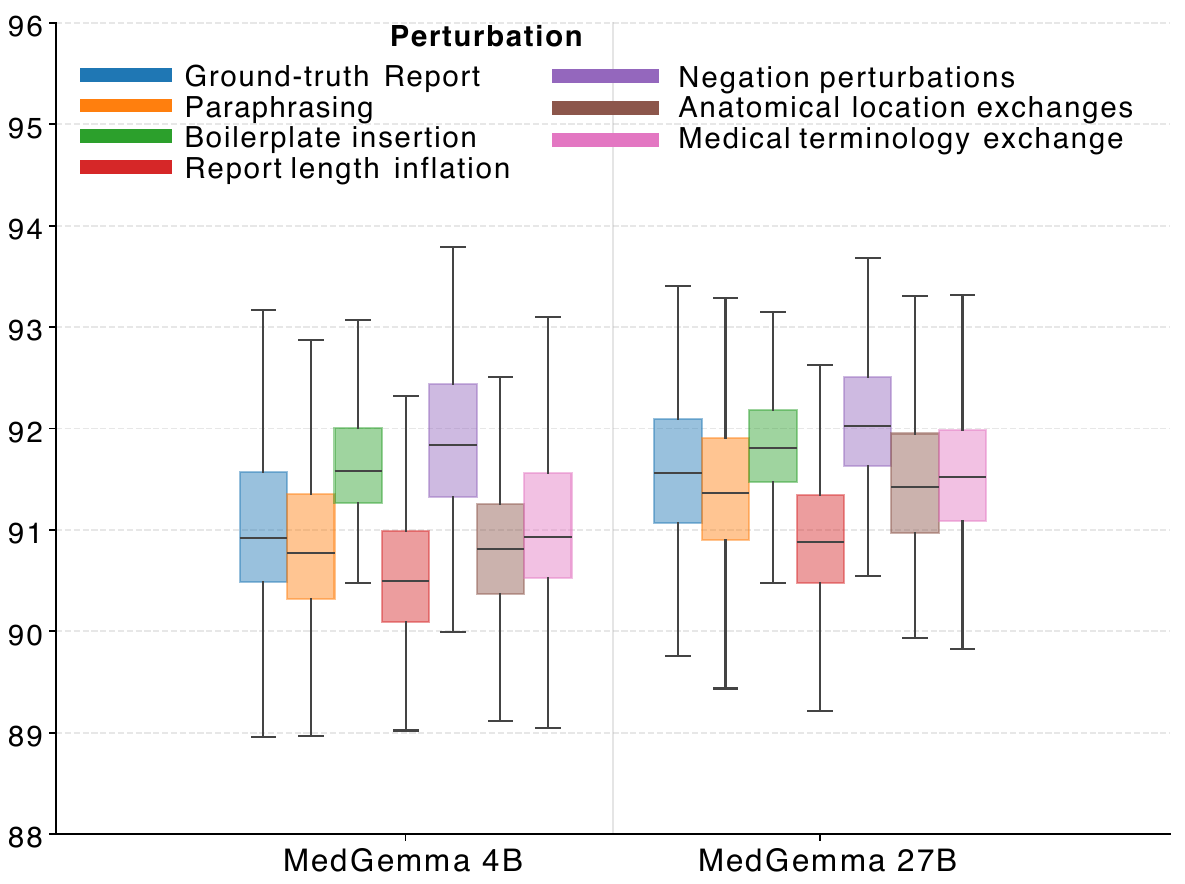}
    \caption{BERTScore sensitivity to six controlled reference-report perturbations for MedGemma-4B and MedGemma-27B, shown relative to the ground-truth baseline.}
    \label{fig:main-metric-perturbations}
\end{figure}

Figure~\ref{fig:main-metric-perturbations} shows semantic and syntactic metric sensitivity to the controlled perturbations.
Paraphrasing caused modest changes (BERTScore $-6.6\%$, BLEU $-14.2\%$),
while superficial structural changes induced larger penalties (BERTScore
$-11.5\%$, BLEU $-12.6\%$). 
Clinically critical perturbations
produced smaller penalties: negation reversals ($-6.6\%$), anatomical
location swaps ($-5.3\%$), and terminology exchanges ($-5.5\%$).
Standard metrics thus penalize stylistic variation more heavily than
diagnostic errors, failing to distinguish correct from
fluent-but-incorrect reports.

\paragraph{IVD-Level Pathology Detection}
Per-level classification (Fig.~\ref{fig:main-classification-findings}),  was near random across all findings and models: 
balanced accuracy ranged 50-57\%, driven by high specificity (74--83\%) and 
low sensitivity (2--34\%), indicating a systematic negative-prediction bias. 
Level stratification revealed a consistent anatomical pattern: near-zero 
sensitivity at upper levels, elevated sensitivity at L4-L5 (70--85\%) with 
sharply reduced specificity (30--40\%), and an analogous pattern at L5-S1 
for MedGemma-27B.

\subsection{Anomaly Detection-Guided Report Generation}
\begin{table}[t]
\centering
\caption{SPIDER anomaly heatmap localization.
DICE and RMSE evaluated on positive ground-truth pixels only;
RMSE after $[0,1]$ normalization. Mean $\pm$ SD over the test set.}
\label{tab:main-anomaly-detection-results}
\resizebox{\columnwidth}{!}{%
\setlength{\tabcolsep}{4pt}
\renewcommand{\arraystretch}{1.15}
\begin{tabular}{l ccccccc}
\toprule
Metric & Overall & T12-L1 & L1-L2 & L2-L3 & L3-L4 & L4-L5 & L5-S1 \\
\midrule
DICE & $82.3\pm2.0$ & $84.1\pm1.9$ & $87.0\pm1.8$ & $82.8\pm2.9$ & $83.8\pm2.4$ & $84.7\pm2.4$ & $79.3\pm3.1$ \\
RMSE & $0.4\pm0.1$  & $0.6\pm0.2$  & $0.4\pm0.2$  & $0.3\pm0.2$  & $0.4\pm0.2$  & $0.1\pm0.1$  & $0.6\pm0.2$  \\
\bottomrule
\end{tabular}%
}
\end{table}

\newpage
\paragraph{Heatmap Localization}

Table~\ref{tab:main-anomaly-detection-results} reports UNet++ heatmap
quality per IVD level. Localization pre-training achieves consistent
disc-level coverage across all six levels, confirming that SpineNetV2
landmarks provide reliable spatial anchors for level-specific heatmap
generation. 
Anomaly fine-tuning improves spatial precision at pathology-specific regions (posterior disc margin, central canal), with gains in DICE and Pearson correlation most pronounced at the clinically prevalent L4-L5 and L5-S1 levels.
Upper levels (T12–L1, L1–L2) have lower absolute scores, reflecting lower pathology prevalence and weaker supervision.
\\

\paragraph{Impact on Report Generation}

\begin{table}[t]
\centering
\caption{Impact of anomaly detection (AD) heatmap overlay on zero-shot IVD-Level Pathology Detection. Shown are sensitivity and specificity across IVDs \textit{img}: central sagittal slice only;
\textit{img\,+\,AD}: slice with heatmap overlay.}
\label{tab:main-anomaly-binary-classification}
\resizebox{\columnwidth}{!}{%
\setlength{\tabcolsep}{6pt}
\renewcommand{\arraystretch}{1.15}
\begin{tabular}{lllcccc}
\toprule
 &  &  & \multicolumn{2}{c}{Sensitivity} & \multicolumn{2}{c}{Specificity} \\
 &  &  & img & img + AD & img & img + AD \\
\midrule
 & \multirow[t]{6}{*}{27B} & T12-L1 & 0.0 & 0.0 & 100.0 & 100.0 \\
 &  & L1-L2 & 2.1 & $\mathbf{22.6}$ & $\mathbf{100.0}$ & 88.8 \\
 &  & L2-L3 & 1.8 & $\mathbf{17.0}$ & $\mathbf{100.0}$ & 88.3 \\
 &  & L3-L4 & 1.5 & $\mathbf{87.5}$ & $\mathbf{100.0}$ & 17.5 \\
 &  & L4-L5 & 73.9 & $\mathbf{100.0}$ & $\mathbf{34.4}$ & 13.4 \\
 &  & L5-S1 & $\mathbf{83.7}$ & 31.3 & 29.2 & $\mathbf{62.9}$ \\
\cline{1-7} 

\bottomrule
\end{tabular}%
}
\end{table}

Adding the heatmap overlay as auxiliary visual input affects report generation metrics compared with the single-image baseline (Sup.~\ref{tab:supp-anomaly-report-generation}).
Models with multi-image conditioning (ChatGPT) consistently improve METEOR and ROUGE-L, indicating that spatially localized anomaly signals promote more anatomically specific language. BERTScore gains are modest, aligning with its known insensitivity to localization precision from the perturbation analysis.
Models without native multi-image support (MedGemma)
show no consistent benefit, confirming that effective integration
requires the model to attend to both inputs jointly. 
Text-guided report generation from AD-based localization can match or surpass zero-shot performance.

\paragraph{Impact on IVD-Level Pathology Detection}
Heatmap integration achieves the largest improvements on the structured 
localization task (Table~\ref{tab:main-anomaly-binary-classification}). Sensitivity
at upper levels (T12-L1 through L3-L4), which is near zero under zero-shot prompting, 
increases for several findings when the overlay highlights anomalous disc 
regions, with specificity largely maintained. This sensitivity gain is 
model- and finding-dependent, suggesting the heatmap provides complementary 
image-derived evidence that partially offsets the anatomical prior bias toward 
lower lumbar levels.

\begin{figure}[t]
  \centering
   \includegraphics[width=0.8\linewidth]{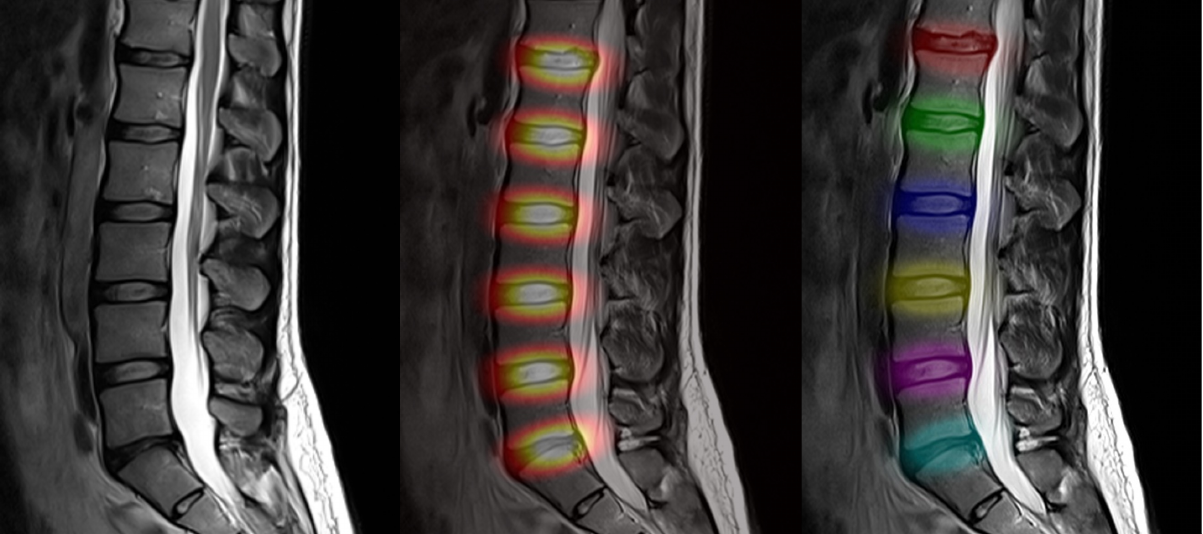}
    \includegraphics[width=0.8\linewidth]{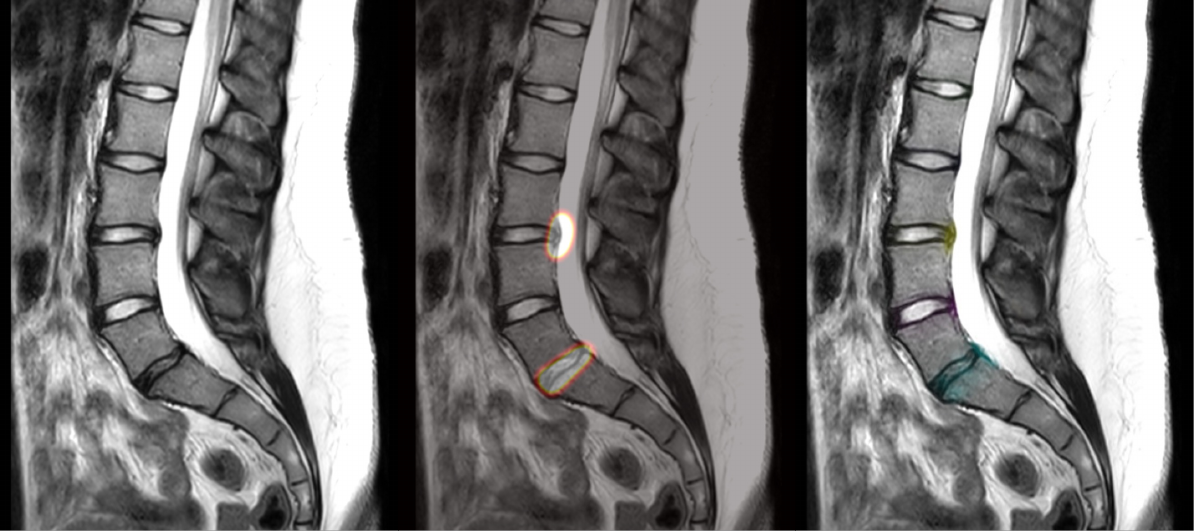}
   \caption{Localization (top) and anomaly (bottom) heatmaps.}
   \label{fig:main-localization-anomaly-heatmaps}
\end{figure}


\subsection{Discussion}

Across experiments, a consistent pattern emerges: leading VLMs
show high semantic similarity but near-random performance on
structured diagnostic tasks, exposing a core gap between
linguistic fluency and MRI-based clinical reasoning.

\noindent\textbf{Fluency does not imply diagnostic correctness.}
MedGemma-27B and ChatGPT-5.0 produce the most fluent reports by all
lexical and semantic metrics, yet their IVD-level pathology detection
remains near random (balanced accuracy 50--56\%). This dissociation
suggest that high BERTScore and METEOR may reflect strong language priors
rather than genuine visual grounding, and that standard language generation metrics are
insufficient proxies for clinical correctness in spine MRI reporting.

\noindent\textbf{Standard metrics show limited sensitivity to clinically important errors.}
Controlled perturbation analysis directly quantifies this failure:
negation reversals, anatomical mislocalization, and terminology swaps
incur smaller metric penalties than boilerplate insertion or
paraphrasing. Therefore, metric-driven optimization  risks rewarding
stylistically fluent but diagnostically incorrect outputs, a silent
failure mode with direct patient safety implications.

\noindent\textbf{Models rely on anatomical priors, not image evidence.}
Input ablations show that changing MRI contrast, resolution, or
prompt formulation produce marginal metric differences,
indicating models are largely invariant to clinically meaningful input
variation. 
IVD-level analysis reveals systematic bias toward
lower lumbar levels (L4-L5, L5-S1), reflecting their higher frequency
in training data rather than image-based reasoning. 

\noindent\textbf{Limited fine-tuning improves metrics but introduces new risks.}
Supervised fine-tuning on the small SPIDER dataset improves aggregate
scores but produces anatomically implausible outputs, such
 as non-existent vertebral levels (e.g., L6–L7), underscoring how metric-driven optimization on limited data can produce text that appears correct but lacks anatomical grounding.
 
\noindent\textbf{Anomaly-guided grounding opens a path forward.}
We propose augmenting VLMs with disc-level anomaly heatmaps as explicit 
auxiliary visual input, designed to improve image-grounded localization. 
By providing spatially localized, image-derived 
evidence at the IVD level, he heatmaps provide spatially localized, image-derived evidence to pathological 
disc regions rather than allowing generation to be driven by language priors alone. 
This results in sensitivity gains at upper lumbar levels (L1–L2), 
where zero-shot models consistently fail, without any architectural 
changes or task-specific VLM retraining.
Importantly, the heatmaps also fulfill an auxiliary function. Anomaly heatmaps  constitute an 
interpretable visual trace that clinicians can evaluate directly, decoupled from 
the generated text. This separation of image-derived evidence from free-text 
generation supports structured quality control and human oversight within 
routine reporting workflows. 
The approach is architecture-agnostic and compatible with 
existing VLMs, 
providing an interpretable and scalable baseline for further study for report generation without reliance on large, densely annotated spine MRI datasets.

\section{Conclusion}


Our findings reveal that leading models produce fluent reports with high semantic similarity scores, while showing near-random performance in structured diagnostic tasks (balanced accuracy 50-56\%). This exposes a critical gap between apparent performance and diagnostic competence, driven by language priors rather than image-grounded reasoning. Standard language-generation metrics amplify this by overlooking clinical errors, such as diagnostic negation or pathology mislocalization, and concomitantly penalizing superficial stylistic variation such as report length or boilerplate phrasing. Spatial grounding relying on anomaly detection offers an architecture-agnostic solution, providing examinable heatmaps that enforce image dependence and increase interpretability, paving the way towards visually-grounded VLMs.

\section*{Acknowledgment}

This project was supported by grant \# 380 of the Strategic Focus Area “Personalized Health and Related Technologies (PHRT)” of the ETH Domain (Swiss Federal Institutes of Technology). 
This research study retrospectively analyzed open access human subject data , exempt from ethical approval according to the open access license.
For the development of this work, AI-assisted coding systems such as Claude Code and Perplexity were used. 
During the preparation of this manuscript, the author(s) reviewed the text for grammar correctness and enhanced the content with the assistance of AI generative tools (Writefull and Perplexity). 
After using this tool/service, the author(s) reviewed and edited the content as needed and take(s) full responsibility for the content of the publication.
The code, along with training and evaluation scripts, and reproducibility instructions, is available at \url{https://gitlab.ethz.ch/BMDSlab/publications/low-back/spine-mri-report-generation}.
{
    \small
    \bibliographystyle{ieeenat_fullname}
    \bibliography{main}
}

\newpage

\clearpage
\setcounter{page}{1}
\maketitlesupplementary


\section{Lumbar MRI Report Generation}

\subsection{Ground Truth Reports for SPIDER Dataset}
\label{app:spider-report-construction}
SPIDER provides structured radiological finding matrices rather than free-text reports. To enable report-generation evaluation, we converted these matrices into synthetic reference reports using predefined sentence templates. This yields standardized yet mildly diverse text that faithfully reflects the annotation matrix.

To reduce bias toward any single phrasing, we introduced limited lexical variation in both anatomical localization and finding description. Level information was expressed using simple constructions such as ``at \{level\}'' or ``\{level\} shows'', while findings were verbalized using a small set of interchangeable formulations, e.g., ``disc space narrowing'' versus ``reduced disc height'', and ``disc bulge'' versus ``disc bulging''.

Only findings annotated as present were included. In contrast to human-written reports, unaffected structures were not described explicitly. When multiple findings co-occurred at the same intervertebral level, they were merged into a single sentence using conjunctions to improve fluency and readability.

Representative templates included:
\begin{itemize}
  \item \textbf{Disc narrowing:} ``At \{LEVEL\}, there is disc space narrowing.'' / ``\{LEVEL\} shows reduced disc height.''
  \item \textbf{Disc bulging:} ``At \{LEVEL\}, there is a disc bulge.'' / ``\{LEVEL\} shows disc bulging.''
  \item \textbf{Disc herniation:} ``At \{LEVEL\}, there is a focal disc herniation.'' / ``\{LEVEL\} shows a focal herniated disc.''
  \item \textbf{Endplate defects:} ``At \{LEVEL\}, there are endplate defects involving the \{upper/lower/both\} endplate(s).''
  \item \textbf{Modic changes:} ``At \{LEVEL\}, Modic-type marrow signal changes are present adjacent to the endplates.''
\end{itemize}

Pfirrmann grading was excluded because it is a composite severity score that subsumes lower-level findings and is typically not stated explicitly in clinical narrative reports. Modic changes were encoded as binary present/absent labels, since subtype differentiation cannot be reliably inferred from T1- or T2-weighted images alone. Endplate defects were described explicitly by anatomical location, distinguishing upper, lower, or combined involvement.

For example, a subject with disc narrowing at L2--L3 and L3--L4, and disc bulging at L3--L4 and L4--L5, would yield:
\begin{quote}
``L2--L3 and L3--L4 show disc space narrowing. L3--L4 and L4--L5 show disc bulging.''
\end{quote}

\subsection{Model-Specific Prompt Configurations}

\label{app:prompt-design}

\begin{table*}
\small
\centering
\caption{Model-specific input configurations across prompt-length settings. Fields marked with $\varnothing$ indicate that no value was passed to the model (i.e., the field was omitted or set to \texttt{None}), whereas entries labeled ``Not specified'' indicate that the field was explicitly provided with that value because it is required by the model interface.}
\label{tab:supp-model-inputs}

\begingroup
\setlength{\parskip}{0pt}

\begin{tabular}{>{\raggedright\arraybackslash}p{0.18\linewidth} >{\raggedright\arraybackslash}p{0.76\linewidth}}
\toprule
\multicolumn{1}{c}{\textbf{Model}} & \multicolumn{1}{c}{\textbf{Input configuration}} \\
\midrule

\multicolumn{2}{l}{\textbf{Short prompt}} \\
\midrule
MedGemma &
\textbf{System role:} You are an expert radiologist.\par
\textbf{Prompt:} Describe the findings for the MRI. \\
\addlinespace[8pt]

ChatGPT &
\textbf{System role:} You are an expert radiologist.\par
\textbf{Prompt:} Describe the findings for the MRI. \\
\addlinespace[8pt]

MAIRA-2 &
\textbf{Indication:} Not specified.\par
\textbf{Technique:} Sagittal MRI of the lumbar spine.\par
\textbf{Comparison:} $\varnothing$.\par
\textbf{Prior frontal image:} $\varnothing$.\par
\textbf{Prior report:} $\varnothing$. \\
\addlinespace[8pt]

VILA-M3 &
\textbf{Prompt:} Describe the findings for the MRI.\par
\textbf{Mode message:} This is a MRI image.\par
\textbf{Expert model cards:} No experts available; answer yourself. \\
\addlinespace[8pt]

BiomedGPT &
\textbf{Task:} Caption. \\

\midrule
\multicolumn{2}{l}{\textbf{Medium prompt}} \\
\midrule
MedGemma &
\textbf{System role:} You are an expert radiologist.\par
\textbf{Prompt:} Describe the findings for the sagittal lumbar spine MRI. \\
\addlinespace[8pt]

ChatGPT &
\textbf{System role:} You are an expert radiologist.\par
\textbf{Prompt:} Describe the findings for the sagittal lumbar spine MRI. \\
\addlinespace[8pt]

MAIRA-2 &
\textbf{Indication:} Lumbar spine degeneration.\par
\textbf{Technique:} Sagittal MRI of the lumbar spine.\par
\textbf{Comparison:} $\varnothing$.\par
\textbf{Prior frontal image:} $\varnothing$.\par
\textbf{Prior report:} $\varnothing$. \\
\addlinespace[8pt]

VILA-M3 &
\textbf{Prompt:} Describe the findings for the sagittal lumbar spine \texttt{\{MRI Modality\}} MRI.\par
\textbf{Mode message:} This is a \texttt{\{MRI Modality\}} MRI image.\par
\textbf{Expert model cards:} No experts available; answer yourself. \\
\addlinespace[8pt]

BiomedGPT &
\textbf{Task:} Caption. \\

\midrule
\multicolumn{2}{l}{\textbf{Long prompt}} \\
\midrule
MedGemma &
\textbf{System role:} You are an expert radiologist.\par
\textbf{Prompt:} Describe the findings and any radiological gradings for this \texttt{\{MRI Modality\}} lumbar spine MRI in a patient with low back pain. \\
\addlinespace[8pt]

ChatGPT &
\textbf{System role:} You are an expert radiologist.\par
\textbf{Prompt:} Describe the findings and any radiological gradings for this \texttt{\{MRI Modality\}} lumbar spine MRI in a patient with low back pain. \\
\addlinespace[8pt]

MAIRA-2 &
\textbf{Indication:} Findings and any radiological gradings in low back pain and lumbar spine degeneration.\par
\textbf{Technique:} Single sagittal \texttt{\{MRI Modality\}} MRI of the lumbar spine.\par
\textbf{Comparison:} $\varnothing$.\par
\textbf{Prior frontal image:} $\varnothing$.\par
\textbf{Prior report:} $\varnothing$. \\
\addlinespace[8pt]

VILA-M3 &
\textbf{Prompt:} Describe the findings and any radiological gradings for this \texttt{\{MRI Modality\}} sagittal lumbar spine MRI in a patient with low back pain.\par
\textbf{Mode message:} This is a \texttt{\{MRI Modality\}} MRI image.\par
\textbf{Expert model cards:} No experts available; answer yourself. \\
\addlinespace[8pt]

BiomedGPT &
\textbf{Task:} Caption. \\

\bottomrule
\end{tabular}
\endgroup
\end{table*}

Different models require different input interfaces, including conversational prompts, task descriptors, and structured metadata fields. To evaluate prompt sensitivity under comparable conditions, we performed a zero-shot prompt-content ablation in which the informational content was held constant across models for each prompt-length setting (short, medium, long), while preserving model-specific formatting requirements.

The exact wording and decomposition of the input followed each model's recommended usage, e.g., system/user role separation for chat-based models and explicit indication or technique fields for structured interfaces. MedGemma, ChatGPT, and VILA-M3 were prompted through conversational inputs, whereas MAIRA-2 and BiomedGPT relied on metadata-style fields rather than role-based prompting.
For models that support free-form report generation from task-defining prompts (MedGemma, ChatGPT, and VILA-M3), we additionally constrained the report boundaries to reduce variability unrelated to clinical content. Specifically, prompts requested a single concise paragraph delimited by \texttt{[[REPORT\_START]]} and \texttt{[[REPORT\_END]]}, enabling deterministic extraction of the report text while minimizing the effect of model-specific preambles or closing statements on downstream evaluation.

\begin{figure*}[h]
    \centering
    \includegraphics[width=0.98\linewidth]{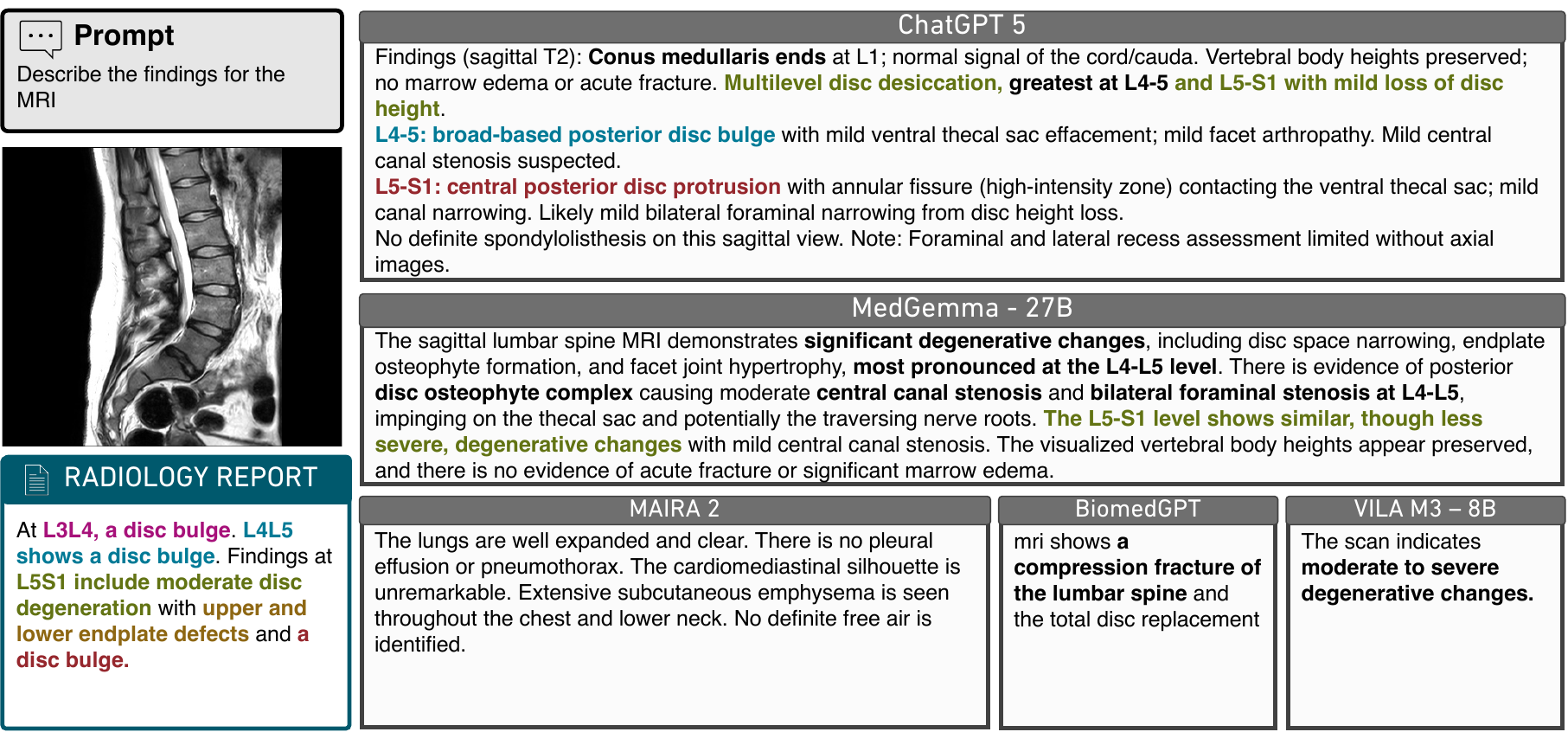}
  \caption{Zero-shot report generation and outputs from identical inputs. Each model (BiomedGPT-Base, ChatGPT-5.0, MAIRA-2, MedGemma-27B, VILA-M3-8B) receives the central sagittal slice, a fixed prompt, and the clinical reference report.}    \label{fig:supp-zero-shot-reports}
\end{figure*}

\begin{figure*}[h]
    \centering
    \includegraphics[width=0.99\linewidth]{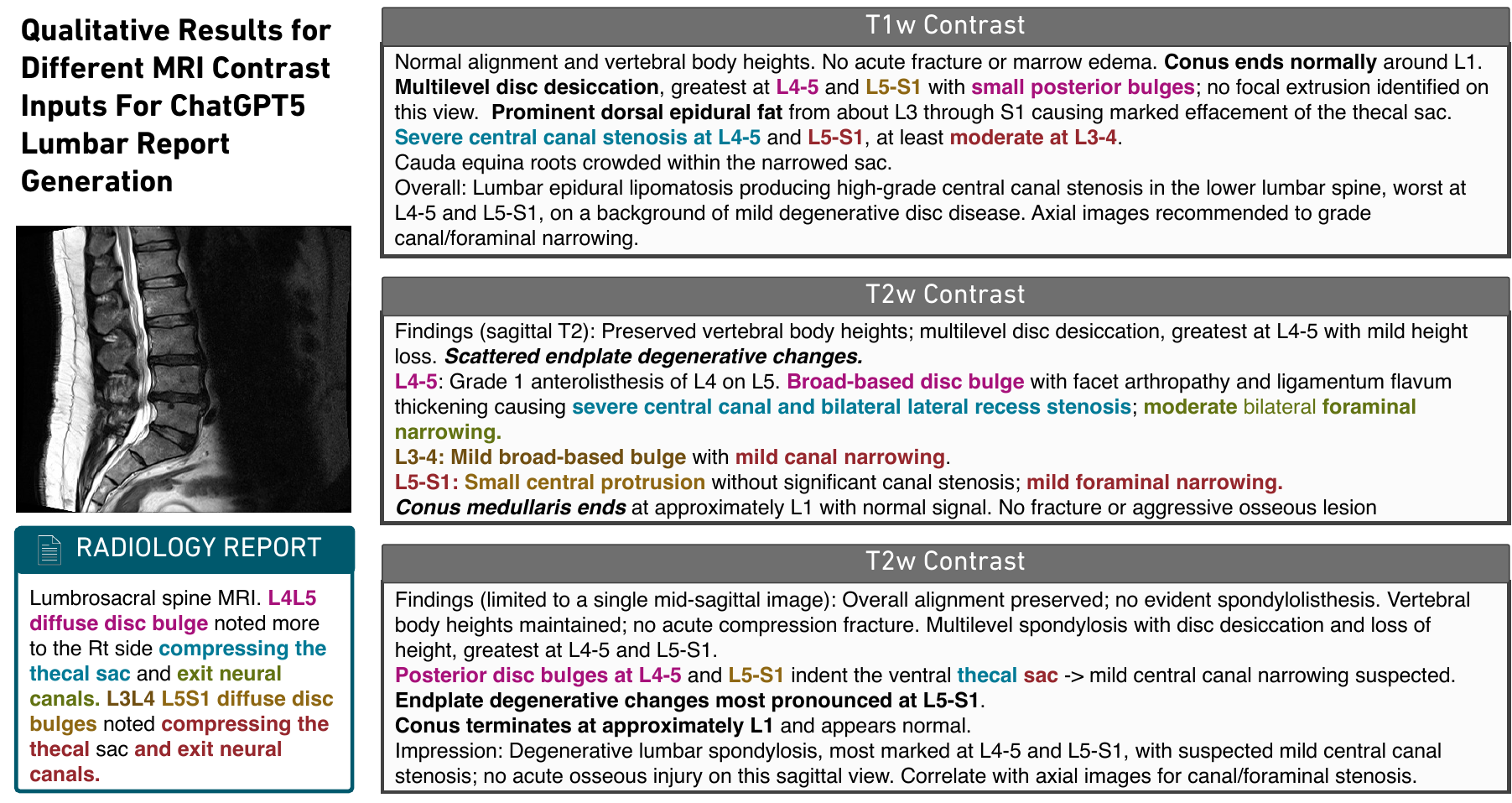}
  \caption{Zero-shot report generation for ChatGPT-5.0 model of the same central slice image for different MRI contrast.}    \label{fig:supp-mri-contrasts}
\end{figure*}

Table~\ref{tab:supp-model-inputs} summarizes the model-specific input configuration used for each prompt-length setting. 
This table specifies the prompt-content ablation, detailing the exact information given to each model under the short, medium, and long settings.
Table~\ref{tab:supp-zero-shot-prompt-ablation} reports the corresponding zero-shot results on LSMRI and SPIDER datasets. Columns correspond to prompt settings (short, medium, long), rows correspond to MRI contrast inputs (T1w, T2w, T1w+T2w), and entries report BERTScore F1, BLEU, METEOR, and ROUGE-L F1, allowing prompt verbosity and image-contrast effects to be compared jointly.
Further qualitative examples illustrating model outputs under different prompt-length and contrast configurations are provided in Fig.~\ref{fig:supp-zero-shot-reports} and Fig.~\ref{fig:supp-mri-contrasts}. 
The results of the central slice variation for the best performing models are summarized further in Table \ref{tab:supp-slice-ablation-results}.

\begin{table*}[tbh]
\caption{Zero-shot report generation prompt-content ablation on LSMRI (free-text reports) and
SPIDER (structured grading reports).
Columns are prompt verbosity settings (Short / Medium / Long) with metrics 
BERTScore~F1 (BERT), BLEU, METEOR, and ROUGE-L~F1.
Rows are MRI contrast inputs (T1w, T2w, T1w+T2w).}
\label{tab:supp-zero-shot-prompt-ablation} 
\centering
\setlength{\tabcolsep}{4pt}
\renewcommand{\arraystretch}{1.15}
\resizebox{0.97\linewidth}{!}{%
\begin{tabular}{ll cccc cccc cccc}


\multicolumn{14}{c}{\textbf{LSMRI — free-text reports}} \\
\toprule
& & \multicolumn{4}{c}{\textbf{Short}} & \multicolumn{4}{c}{\textbf{Medium}} & \multicolumn{4}{c}{\textbf{Long}} \\
\cmidrule(lr){3-6}\cmidrule(lr){7-10}\cmidrule(lr){11-14}

Model & Contrast &
BERT & BLEU & METEOR & ROUGE-L &
BERT & BLEU & METEOR & ROUGE-L &
BERT & BLEU & METEOR & ROUGE-L \\
\toprule

\multirow{3}{*}{ChatGPT-5.0}
& T1w     & 0.913 & 0.005 & 0.226 & 0.122 & 0.911 & 0.005 & 0.214 & 0.117 & 0.901 & 0.001 & 0.137 & 0.074 \\
& T2w     & 0.913 & 0.005 & 0.217 & 0.120 & 0.911 & 0.005 & 0.213 & 0.116 & 0.901 & 0.001 & 0.139 & 0.071 \\
& T1w+T2w & 0.913 & 0.005 & 0.232 & 0.129 & 0.911 & 0.004 & 0.215 & 0.117 & 0.901 & 0.001 & 0.146 & 0.073 \\
\midrule

\multirow{3}{*}{MAIRA-2}
& T1w     & 0.894 & 0.001 & 0.064 & 0.071 & 0.896 & 0.001 & 0.070 & 0.073 & 0.897 & 0.002 & 0.073 & 0.079 \\
& T2w     & 0.895 & 0.002 & 0.069 & 0.069 & 0.897 & 0.002 & 0.075 & 0.075 & 0.894 & 0.001 & 0.068 & 0.080 \\
& T1w+T2w & 0.890 & 0.001 & 0.052 & 0.085 & 0.890 & 0.001 & 0.053 & 0.085 & 0.897 & 0.001 & 0.082 & 0.076 \\
\midrule

\multirow{3}{*}{MedGemma-4B}
& T1w     & 0.908 & 0.006 & 0.127 & 0.127 & 0.911 & 0.007 & 0.142 & 0.143 & 0.911 & 0.008 & 0.157 & 0.141 \\
& T2w     & 0.909 & 0.007 & 0.133 & 0.130 & 0.911 & 0.005 & 0.149 & 0.152 & 0.912 & 0.006 & 0.159 & 0.149 \\
& T1w+T2w & 0.908 & 0.005 & 0.128 & 0.130 & 0.912 & 0.006 & 0.153 & 0.159 & 0.910 & 0.008 & 0.169 & 0.156 \\
\midrule

\multirow{3}{*}{MedGemma-27B}
& T1w     & 0.918 & 0.013 & 0.249 & 0.155 & 0.916 & 0.011 & 0.227 & 0.163 & 0.911 & 0.007 & 0.199 & 0.117 \\
& T2w     & 0.917 & 0.012 & 0.239 & 0.150 & 0.916 & 0.013 & 0.230 & 0.164 & 0.916 & 0.008 & 0.230 & 0.139 \\
& T1w+T2w & 0.911 & 0.010 & 0.194 & 0.138 & 0.915 & 0.010 & 0.227 & 0.167 & 0.909 & 0.009 & 0.203 & 0.115 \\
\midrule

\multirow{3}{*}{VILA-M3 3B}
& T1w     & 0.890 & 0.001 & 0.043 & 0.048 & 0.890 & 0.000 & 0.038 & 0.045 & 0.889 & 0.000 & 0.055 & 0.048 \\
& T2w     & 0.891 & 0.001 & 0.045 & 0.049 & 0.889 & 0.000 & 0.051 & 0.048 & 0.889 & 0.000 & 0.055 & 0.048 \\
& T1w+T2w & 0.895 & 0.001 & 0.056 & 0.065 & 0.891 & 0.001 & 0.059 & 0.063 & 0.889 & 0.000 & 0.055 & 0.048 \\
\midrule

\multirow{3}{*}{VILA-M3 8B}
& T1w     & 0.890 & 0.001 & 0.038 & 0.045 & 0.907 & 0.001 & 0.033 & 0.053 & 0.902 & 0.001 & 0.038 & 0.046 \\
& T2w     & 0.892 & 0.001 & 0.035 & 0.040 & 0.908 & 0.000 & 0.027 & 0.045 & 0.902 & 0.001 & 0.041 & 0.047 \\
& T1w+T2w & 0.898 & 0.001 & 0.033 & 0.039 & 0.894 & 0.000 & 0.034 & 0.041 & 0.898 & 0.001 & 0.057 & 0.055 \\
\midrule

\multirow{3}{*}{VILA-M3 13B}
& T1w     & 0.874 & 0.000 & 0.024 & 0.039 & 0.872 & 0.001 & 0.024 & 0.039 & 0.887 & 0.000 & 0.027 & 0.041 \\
& T2w     & 0.874 & 0.000 & 0.024 & 0.039 & 0.873 & 0.000 & 0.023 & 0.039 & 0.884 & 0.001 & 0.029 & 0.049 \\
& T1w+T2w & 0.880 & 0.000 & 0.021 & 0.043 & 0.886 & 0.001 & 0.026 & 0.048 & 0.895 & 0.000 & 0.041 & 0.047 \\
\bottomrule

\multicolumn{14}{c}{ } \\

\multicolumn{14}{c}{\textbf{SPIDER - structured grading reports}} \\
\toprule
& & \multicolumn{4}{c}{\textbf{Short}} & \multicolumn{4}{c}{\textbf{Medium}} & \multicolumn{4}{c}{\textbf{Long}} \\
\cmidrule(lr){3-6}\cmidrule(lr){7-10}\cmidrule(lr){11-14}
Model & Contrast &
BERT & BLEU & METEOR & ROUGE-L &
BERT & BLEU & METEOR & ROUGE-L &
BERT & BLEU & METEOR & ROUGE-L \\
\toprule
\multirow{3}{*}{ChatGPT-5.0}
& T1w     & 0.922 & 0.009 & 0.183 & 0.164 & 0.922 & 0.009 & 0.191 & 0.166 & 0.920 & 0.010 & 0.235 & 0.161 \\
& T2w     & 0.922 & 0.008 & 0.191 & 0.165 & 0.922 & 0.009 & 0.198 & 0.169 & 0.919 & 0.008 & 0.230 & 0.155 \\
& T1w+T2w & 0.922 & 0.004 & 0.192 & 0.166 & 0.919 & 0.006 & 0.189 & 0.159 & 0.918 & 0.009 & 0.235 & 0.155 \\
\midrule

\multirow{3}{*}{MAIRA-2}
& T1w     & 0.899 & 0.007 & 0.084 & 0.066 & 0.905 & 0.009 & 0.089 & 0.073 & 0.904 & 0.012 & 0.091 & 0.075 \\
& T2w     & 0.899 & 0.006 & 0.085 & 0.064 & 0.905 & 0.009 & 0.087 & 0.071 & 0.903 & 0.011 & 0.090 & 0.072 \\
& T1w+T2w & 0.897 & 0.001 & 0.066 & 0.052 & 0.900 & 0.006 & 0.079 & 0.062 & 0.903 & 0.008 & 0.096 & 0.069 \\
\midrule

\multirow{3}{*}{MedGemma-4B}
& T1w     & 0.916 & 0.004 & 0.099 & 0.088 & 0.918 & 0.004 & 0.098 & 0.093 & 0.918 & 0.011 & 0.113 & 0.101 \\
& T2w     & 0.917 & 0.005 & 0.102 & 0.092 & 0.918 & 0.004 & 0.096 & 0.094 & 0.920 & 0.010 & 0.116 & 0.111 \\
& T1w+T2w & 0.914 & 0.003 & 0.088 & 0.086 & 0.917 & 0.003 & 0.088 & 0.088 & 0.916 & 0.011 & 0.105 & 0.100 \\
\midrule

\multirow{3}{*}{MedGemma-27B}
& T1w     & 0.927 & 0.029 & 0.202 & 0.160 & 0.925 & 0.026 & 0.184 & 0.151 & 0.928 & 0.033 & 0.226 & 0.154 \\
& T2w     & 0.927 & 0.027 & 0.196 & 0.159 & 0.923 & 0.024 & 0.172 & 0.140 & 0.931 & 0.041 & 0.234 & 0.183 \\
& T1w+T2w & 0.918 & 0.019 & 0.145 & 0.122 & 0.917 & 0.015 & 0.135 & 0.114 & 0.920 & 0.022 & 0.190 & 0.135 \\
\midrule

\multirow{3}{*}{VILA-M3 3B}
& T1w     & 0.898 & 0.009 & 0.065 & 0.036 & 0.899 & 0.016 & 0.080 & 0.050 & 0.899 & 0.017 & 0.085 & 0.053 \\
& T2w     & 0.899 & 0.010 & 0.068 & 0.040 & 0.899 & 0.017 & 0.084 & 0.053 & 0.899 & 0.017 & 0.086 & 0.054 \\
& T1w+T2w & 0.904 & 0.004 & 0.048 & 0.044 & 0.899 & 0.013 & 0.078 & 0.055 & 0.899 & 0.017 & 0.086 & 0.054 \\
\midrule

\multirow{3}{*}{VILA-M3 8B}
& T1w     & 0.899 & 0.009 & 0.059 & 0.037 & 0.911 & 0.000 & 0.025 & 0.029 & 0.906 & 0.000 & 0.030 & 0.023 \\
& T2w     & 0.900 & 0.007 & 0.055 & 0.034 & 0.914 & 0.000 & 0.027 & 0.037 & 0.906 & 0.000 & 0.029 & 0.018 \\
& T1w+T2w & 0.900 & 0.000 & 0.028 & 0.015 & 0.902 & 0.000 & 0.025 & 0.011 & 0.903 & 0.001 & 0.042 & 0.026 \\
\midrule

\multirow{3}{*}{VILA-M3 13B}
& T1w     & 0.878 & 0.000 & 0.017 & 0.001 & 0.884 & 0.002 & 0.029 & 0.034 & 0.883 & 0.000 & 0.023 & 0.016 \\
& T2w     & 0.878 & 0.000 & 0.017 & 0.000 & 0.881 & 0.000 & 0.026 & 0.036 & 0.887 & 0.001 & 0.028 & 0.035 \\
& T1w+T2w & 0.879 & 0.000 & 0.016 & 0.009 & 0.884 & 0.000 & 0.016 & 0.021 & 0.884 & 0.000 & 0.026 & 0.036 \\
\bottomrule
\end{tabular}%
}
\end{table*}

\begin{table*}[bht]
\centering
\caption{Zero-shot report generation performance under different slice selection configurations on LSMRI and SPIDER. Columns are metrics (BERTScore F1, BLEU, METEOR, ROUGE-L F1). Rows are slice index selected for input with respect to the central index.}
\resizebox{\linewidth}{!}{%
\setlength{\tabcolsep}{4pt}
\begin{tabular}{lllcccccccc}
\toprule
\multicolumn{3}{c}{} &
\multicolumn{4}{c}{\textbf{LSMRI  Dataset (Free Text Report)}} &
\multicolumn{4}{c}{\textbf{SPIDER Dataset (Structured Grading Report)}} \\
\cmidrule(lr){4-7}\cmidrule(lr){8-11}
Model & Size & Stack &
BERTScore F1 & BLEU & METEOR & ROUGE-L F1 &
BERTScore F1 & BLEU & METEOR & ROUGE-L F1 \\
\midrule

\multirow{6}{*}{MedGemma}
  & \multirow{3}{*}{4B}  & Central Slice & 0.911 & 0.005 & 0.149 & 0.152 & 0.918 & 0.004 & 0.094 & 0.092 \\
  &                      &\quad $\pm 1$       & 0.912 & 0.007 & 0.153 & 0.153 & 0.919 & 0.005 & 0.099 & 0.096 \\
  &                      & $\quad \pm 2$       & 0.911 & 0.007 & 0.151 & 0.151 & 0.919 & 0.004 & 0.101 & 0.098 \\
\cline{2-11}
  & \multirow{3}{*}{27B} & Central Slice & 0.916 & 0.013 & 0.230 & 0.164 & 0.922 & 0.022 & 0.164 & 0.135 \\
  &                      & \quad $\pm 1$       & 0.916 & 0.013 & 0.228 & 0.163 & 0.923 & 0.025 & 0.169 & 0.137 \\
  &                      &  \quad $\pm 2$       & 0.915 & 0.011 & 0.218 & 0.159 & 0.924 & 0.024 & 0.176 & 0.140 \\
\bottomrule
\end{tabular}
}
\label{tab:supp-slice-ablation-results}
\end{table*}


\subsection{Fine-tuning Implementation Details}
\label{app:finetuning}

\begin{figure*}[htb]
    \centering
    \includegraphics[width=0.99\linewidth]{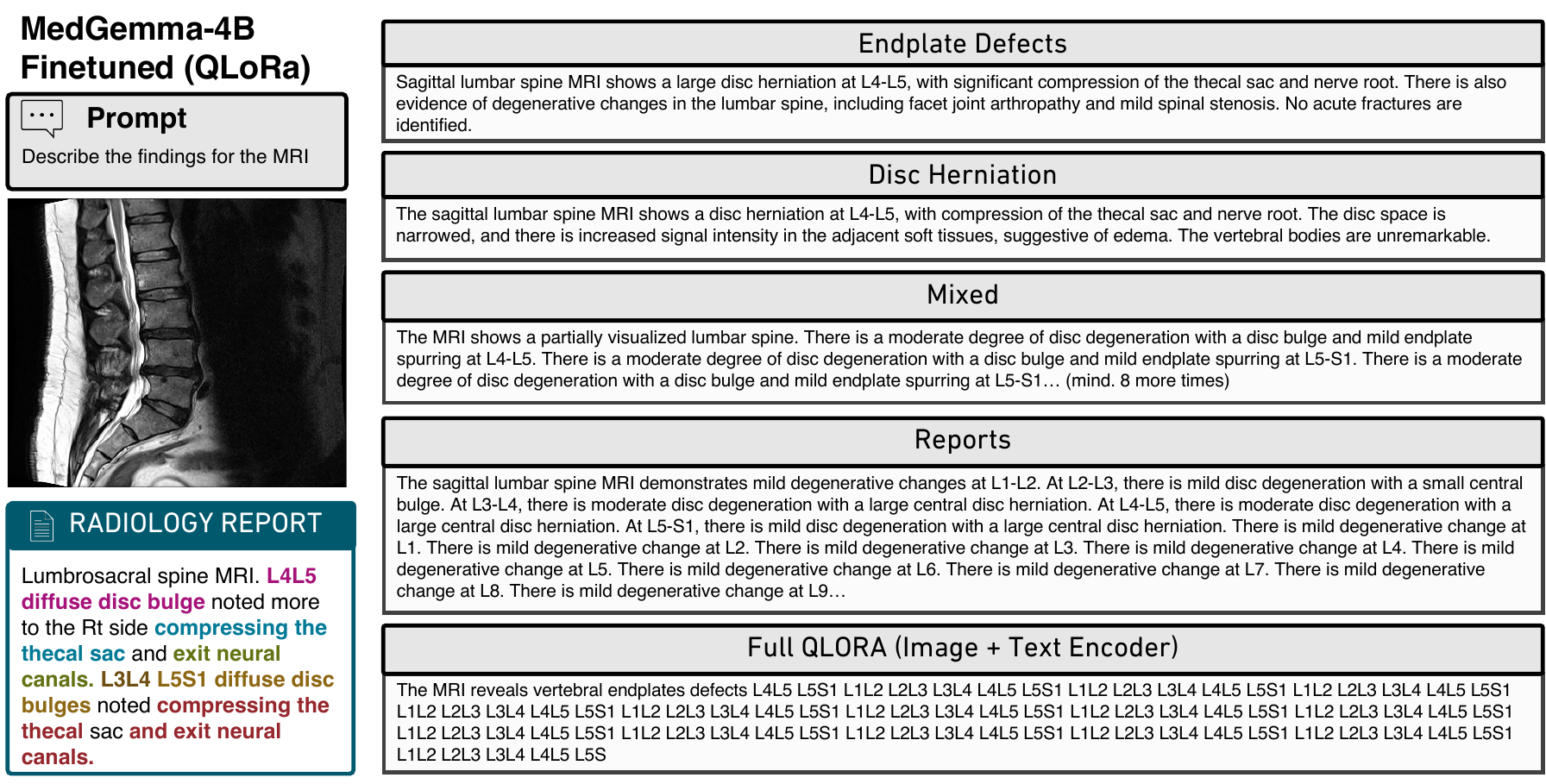}
    \caption{Qualitative comparison of MedGemma-4B outputs after fine-tuning with different SPIDER-derived supervision targets shows that finding-specific, mixed, and report-based supervision produce distinct output structures.}
    \label{fig:supp-finetuning-examples}
\end{figure*}

We fine-tuned MedGemma-4B with QLoRA, updating 20.4M of 4.32B parameters (0.47\%) while keeping the pretrained base weights frozen in 4-bit NF4 format and optimizing LoRA adapters in bfloat16 precision. 
Optimization used AdamW with cosine learning-rate decay, 3\% linear warmup, gradient accumulation of 16 steps, and the selected hyperparameters \(lr = 2^{-4}\), \(r = 32\), and \(\alpha = 64\); models were trained with an 80/20 patient-level train/validation split and selected by best validation loss.  
Training used SPIDER supervision converted into either grading-derived report text or per-finding level targets, with both T1w and T2w sagittal MRI inputs.  
Figure~\ref{fig:supp-finetuning-examples} shows a representative qualitative example of the fine-tuning strategy.

By \emph{fine-tuning strategy}, we mean both the parameter-efficient adaptation setup and the supervision method used during training.
We evaluated several supervision regimes derived from SPIDER, including: separate finding-specific targets (e.g., disc bulging, disc herniation, disc narrowing, endplate defects, Modic changes, Pfirrmann grade, spondylolisthesis), a mixed-finding subset, and full report-style targets.
This distinction is important, as the fine-tuned variants differ mainly in their training label space, while the optimization framework remained fixed across experiments.

\section{Metric Robustness Analysis for Clinical Application}
\label{app:metric-perturbations}

For the metric robustness analysis, we constructed synthetic report augmentations by applying deterministic, rule-based transformations to the original reference reports.
Each augmentation category was designed to isolate a specific lexical or clinical-semantic change while preserving the original report structure whenever possible.
All perturbations were applied independently, and each transformed report was compared against its unmodified reference.

\subsection{Reference Conditions}

\paragraph{Exact Match}
The exact match condition was used as the reference corresponding to a normalized score of 100 for each evaluation metric. In this setting, each report was evaluated against itself.

\paragraph{Random Sampling}
The random sampling condition was used as the reference corresponding to a normalized score of 0 for each evaluation metric. For this baseline, each report was evaluated against a synthetic report of identical length generated by uniformly sampling words from the empirical word distribution of the full report corpus.

\subsection{Experiment Conditions}
\begin{figure*}[th]
    \centering
    \includegraphics[width=0.8\linewidth]{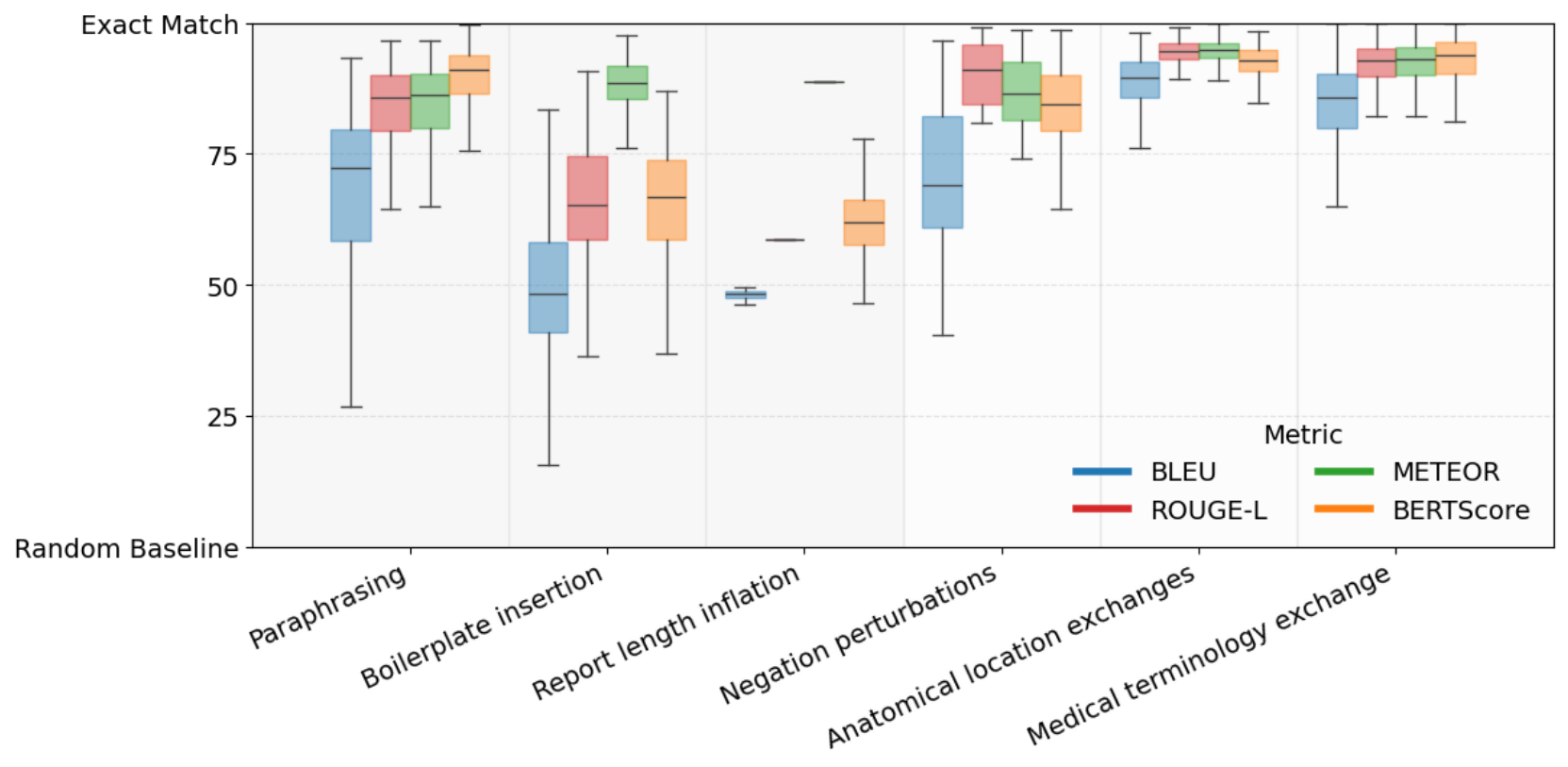}
    \caption{Distribution of normalized report-generation scores under six controlled perturbations to reference reports only. Surface-form changes such as boilerplate insertion and length inflation cause large score shifts, while several clinically important perturbations receive only small penalties.}
    \label{fig:supp-metric-robustness}
\end{figure*}
\paragraph{Negation Perturbations}
Negation perturbations were generated by systematically removing or replacing negation expressions that reverse diagnostic polarity. Common negation tokens such as \emph{no}, \emph{not}, \emph{without}, \emph{neither}, \emph{nor}, \emph{never}, \emph{absence}, and \emph{negative} were removed or substituted using deterministic, rule-based transformations.
In cases where simple removal of negation terms resulted in grammatically incorrect or clinically implausible sentences, phrase-level substitutions were applied to ensure syntactic correctness and semantic consistency. These substitutions were defined manually based on frequent patterns observed in the reference reports and were applied using case-insensitive exact matching. Representative examples are shown in Table~\ref{tab:supp-negation-substitutions}.

\begin{table}[h]
\centering
\caption{Representative phrase-level substitutions applied during negation perturbations.}
\label{tab:supp-negation-substitutions}
\begin{tabular}{p{0.45\linewidth} p{0.45\linewidth}}
\toprule
Original phrase & Replacement phrase \\
\midrule
no significant thecal sac or nerve root compression noted & significant thecal sac and nerve root compression noted \\
no disc protrusion or herniation noted & disc protrusion and herniation noted \\
no disc herniation or protrusion noted & disc protrusion and herniation noted \\
no evidence of & evidence of \\
negative for & positive for \\
no significant & significant \\
\bottomrule
\end{tabular}
\end{table}

\paragraph{Anatomical Location Exchanges}
Anatomical location exchanges were generated by substituting vertebral levels or regional descriptors with plausible but incorrect alternatives while preserving sentence structure. Representative substitutions are listed in Table~\ref{tab:supp-location-substitutions}.

\begin{table}[h]
\centering
\caption{Representative anatomical location substitutions applied during location exchange perturbations.}
\label{tab:supp-location-substitutions}
\begin{tabular}{p{0.4\linewidth} p{0.4\linewidth}}
\toprule
Original term & Replacement term \\
\midrule
L1--L2 & L2--L3 \\
L2--L3 & L3--L4 \\
L3--L4 & L4--L5 \\
L4--L5 & L5--S1 \\
L5--S1 & L4--L5 \\
\midrule
lumbar & thoracic \\
thoracic & lumbar \\
sacral & lumbar \\
\bottomrule
\end{tabular}
\end{table}

\paragraph{Paraphrasing}
Paraphrasing perturbations were generated using lexical substitutions that preserve semantic meaning while altering surface form. Representative examples are listed in Table~\ref{tab:supp-paraphrasing-substitutions}.

\begin{table}[h]
\centering
\caption{Representative paraphrasing substitutions applied during paraphrasing perturbations.}
\label{tab:supp-paraphrasing-substitutions}
\begin{tabular}{p{0.45\linewidth} p{0.45\linewidth}}
\toprule
Original term or phrase & Replacement term or phrase \\
\midrule
maintained & preserved \\
noted & observed \\
small & minor \\
mild & slight \\
compatible with & consistent with \\
various & several \\
largely & mostly \\
\bottomrule
\end{tabular}
\end{table}

\paragraph{Medical Terminology Exchange}
Medical terminology exchange perturbations were generated by replacing semantically equivalent medical terms commonly used interchangeably in radiology reports. These substitutions preserve diagnostic meaning while altering surface form. Representative examples are listed in Table~\ref{tab:supp-medical-terminology-exchange}.

\begin{table}[h]
\centering
\caption{Representative medical terminology substitutions applied during terminology exchange perturbations.}
\label{tab:supp-medical-terminology-exchange}
\begin{tabular}{p{0.45\linewidth} p{0.45\linewidth}}
\toprule
Original term & Replacement term \\
\midrule
disc bulge & disc herniation \\
disc herniation & disc bulge \\
stenosis & narrowing \\
narrowing & stenosis \\
neural foraminal narrowing & foraminal stenosis \\
spondylolisthesis & vertebral slippage \\
vertebral slippage & spondylolisthesis \\
\bottomrule
\end{tabular}
\end{table}

\paragraph{Boilerplate Insertion}
Boilerplate insertion perturbations were generated by inserting standardized preambles or closing statements commonly produced by large language models. These insertions do not replace existing content but introduce additional templated structure and verbosity. Representative examples are shown in Table~\ref{tab:supp-boilerplate-insertions}.

\begin{table}[h]
\centering
\caption{Representative boilerplate phrases inserted during boilerplate insertion perturbations.}
\label{tab:supp-boilerplate-insertions}
\begin{tabular}{p{0.25\linewidth} p{0.65\linewidth}}
\toprule
Insertion position & Inserted phrase \\
\midrule
Prefix & Certainly, here is the report based on the MRI image: \\
Prefix & Below is the generated report based on the MRI scan: \\
Suffix & This concludes the MRI report. \\
Suffix & No additional significant abnormalities are identified. \\
\bottomrule
\end{tabular}
\end{table}

\paragraph{Report Length Inflation}
Report length inflation was generated by duplicating each report in its entirety and concatenating the two copies, resulting in a report containing the same words twice in the original order. This augmentation preserves lexical content and semantic meaning while increasing verbosity and repetition.

\subsection{Metric Behavior on Perturbed Reference Reports}

To isolate metric behavior from model performance, this analysis was performed on the reference reports alone. Each experiment compares an original report with a deterministically perturbed version of the same report, without using model-generated text. Figure~\ref{fig:supp-metric-robustness} summarizes the resulting score distributions across perturbation types and highlights the differing sensitivity of standard metrics to lexical versus clinically meaningful changes.

\subsection{IVD-level Binary Classification of Degenerative Findings}
\label{app:per-level-label-generation}

We derived per-level classification targets from the structured radiological grading matrices provided in SPIDER and LumbarDISC. For each finding, the target output consists of the set of intervertebral disc (IVD) levels at which the abnormality is present.
Pfirrmann grading and vertebral levels outside the T12--S1 range were excluded from this task. The same label-generation procedure was applied to LumbarDISC, restricted to spinal canal stenosis, which was the only finding used in our per-level evaluation.

Given the radiological gradings (Tab.~\ref{tab:supp-spider-example-matrix}), we generate one target string per finding by enumerating all positive IVD levels. 
For example:
\begin{itemize}
    \item \textbf{Disc narrowing}: \texttt{T12L1 L1L2 L2L3 L3L4 L5S1}
    \item \textbf{Spondylolisthesis}: \texttt{None}
    \item \textbf{Disc bulging}: \texttt{L2L3 L3L4 L4L5 L5S1}
\end{itemize}

\begin{table}[bht]
\centering
\caption{Example SPIDER  radiological grading matrix. Binary values denote absence (0) or presence (1), except endplate defects, which encode location: 0 = absent, 1 = upper endplate only, 2 = lower endplate only, 3 = both endplates.}
\label{tab:supp-spider-example-matrix}
\resizebox{\columnwidth}{!}{%
\setlength{\tabcolsep}{4pt}
\renewcommand{\arraystretch}{1.1}
\begin{tabular}{lcccccc}
\toprule
Finding & T12--L1 & L1--L2 & L2--L3 & L3--L4 & L4--L5 & L5--S1 \\
\midrule
Disc narrowing     & 1 & 1 & 1 & 1 & 0 & 1 \\
Spondylolisthesis  & 0 & 0 & 0 & 0 & 0 & 0 \\
Endplate defects   & 2 & 0 & 0 & 0 & 0 & 0 \\
Disc bulging       & 0 & 0 & 1 & 1 & 1 & 1 \\
Disc herniation    & 0 & 0 & 0 & 0 & 0 & 0 \\
Modic changes      & 0 & 0 & 0 & 0 & 0 & 0 \\
\bottomrule
\end{tabular}%
}
\end{table}
\newpage
Each finding-specific label was independently predicted using the following standardized prompt: \newline

\begin{samepage}
\noindent\fbox{%
  \parbox{\columnwidth}{%
    \footnotesize\ttfamily
    Target finding: \{FINDING\}.\\[2pt]
    Valid spinal levels to choose from (use ONLY these, and only if clearly present in the image): \{LEVELS\}.\\[6pt]

    OUTPUT RULES (STRICT):\\
    - If the target finding is present: output ONLY the matching levels from the list above, separated by single spaces (e.g., L3L4 L4L5).\\
    - If the target finding is NOT present (or is ambiguous/uncertain): output EXACTLY None.\\
    - Use UPPERCASE exactly as shown. No extra text, punctuation, labels, or newlines. One line only.\\[6pt]

    Now examine the MRI and output either the space-separated levels or None.
  }%
}
\end{samepage}

\section{Anomaly-Guided Report Generation}

\subsection{Weakly-Supervised Heatmap Generation}
\label{app:pseudo-anomaly-heatmaps}

Because pixel-level pathology annotations are unavailable for lumbar spine MRI
at scale, we derive weakly supervised spatial targets from the structured
per-IVD grading labels provided by the SPIDER dataset.
For each annotated finding and each positive IVD level, we place an
anatomically motivated coarse region within the corresponding channel of the
six-channel heatmap tensor $\mathbf{H} \in \mathbb{R}^{6 \times H \times W}$,
following the expected spatial distribution of that pathology.
Specifically:

\begin{itemize}
    \item \textbf{Disc bulge / herniation:} shallow posterior caps at the posterior disc margin, clipped to retain only posterior-facing support, modeling spinal canal protrusion.
    \item \textbf{Spinal canal stenosis:} larger posterior regions extending toward the expected canal location.
    \item \textbf{Disc narrowing:} contracted quadrilateral regions spanning the entire disc space.
    \item \textbf{Endplate defects:} thin bands along the superior or inferior endplate of adjacent vertebral bodies, coded by location (upper, lower, or both).
    \item \textbf{Modic changes:} trapezoidal regions extending from the endplates into the subchondral vertebral marrow.
    \item \textbf{Spondylolisthesis:} a continuous posterior displacement trace across the disc space and adjacent vertebral bodies.
\end{itemize}

When multiple findings co-occur at the same IVD level, their
finding heatmaps are combined by a per-pixel maximum to prevent
artificial intensity accumulation.
All targets are modeled as soft spatial distributions using 
anisotropic Gaussian smoothing aligned with the estimated disc orientation, allowing for imprecise landmark estimates from SpineNetV2~\cite{windsor_spinenetv2_2022}.
Per-finding and per-channel normalization balances supervision across findings with different spatial extents and disc levels with different prevalences.

\begin{figure}[t]
    \centering
    \includegraphics[width=1\linewidth]{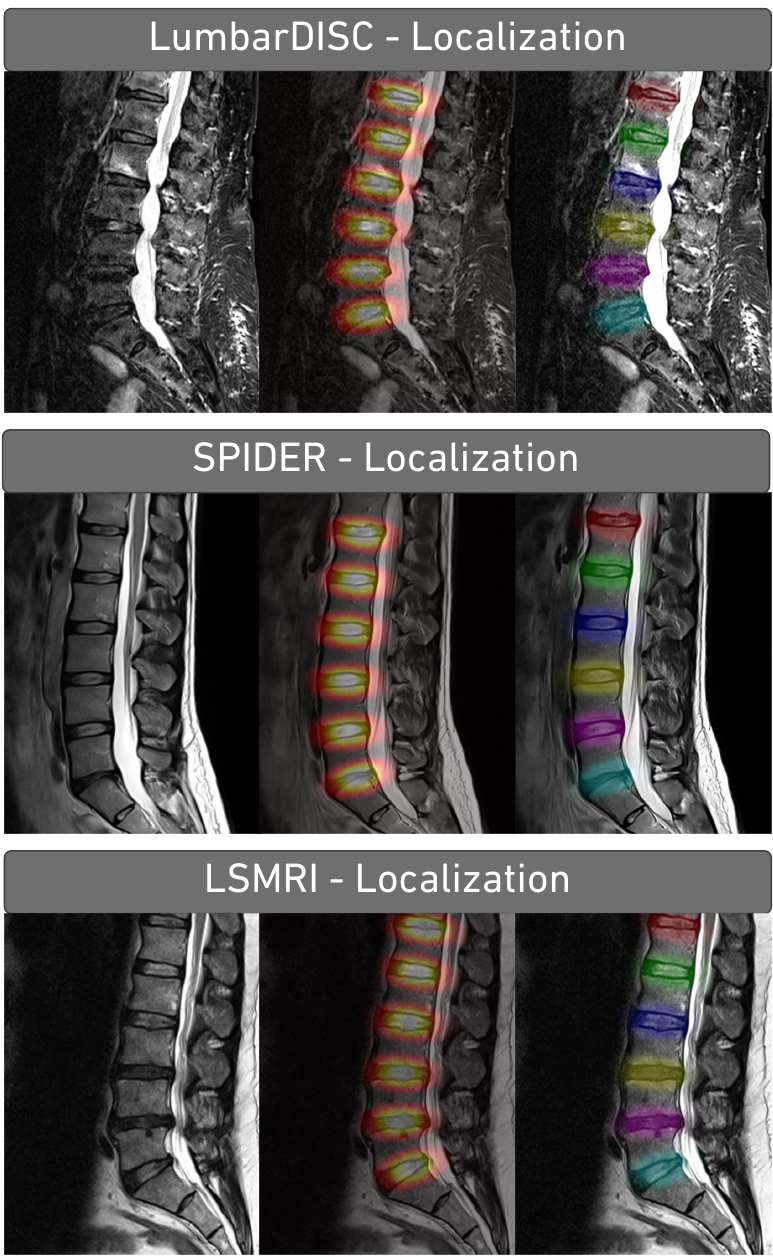}
    \caption{
    \textbf{Localization heatmaps test examples of each dataset.}
    \emph{Left}: central sagittal T2-weighted MRI slice.
    \emph{Center}: ground truth six-channel localization heatmap.
   \emph{Right}: predicted alpha-blended overlay of localization heatmaps.}
    \label{fig:supp-localization-examples}
\end{figure}

\subsection{Anomaly Detection Training}
Anomaly detection training is done in two phases, localization and anomaly detection.
In the first phase, the model is pre-trained exclusively on disc-level
\emph{localization} heatmaps $\mathbf{H}^\text{loc}$ (coarse quadrilateral
disc regions) on 85\% of LumbarDISC and 30\% of LSMRI to establish
level-consistent disc localization.
In the second phase, supervision is gradually shifted toward
\emph{anomaly} heatmaps $\mathbf{H}^\text{AD}$ via linear interpolation:
\begin{equation}
    \mathbf{H}^t = (1 - t)\,\mathbf{H}^\text{loc} + t\,\mathbf{H}^\text{AD},
    \quad t \in [0,1],
    \label{eq:supp-curriculum}
\end{equation}
where $t$ increases monotonically over training epochs.
A channel-exclusivity penalty $\mathcal{L}_\text{IVD}$ is ramped up concurrently to discourage co-activation across IVD channels.

\subsubsection{Training data splits}

\paragraph{Localization Pre-Training} The U-Net is pre-trained using disc-level localization heatmaps $\mathbf{H}^\text{loc}$ on 85\% of LumbarDISC ($n\approx1{,}679$) and 30\% of LSMRI ($n\approx155$),  for 351 epochs with a fixed learning rate of $10^{-3}$.
\paragraph{Anomaly Fine-Tuning} The model is fine-tuned using  pathology-specific heatmaps $\mathbf{H}^\text{AD}$ derived from SPIDER ($n=218$, 80/10/10 patient-level train/val/test split)  for 501 epochs, with the supervision schedule of Eq.~(2) ramping  $t$ linearly from 0 to 1 over the first 1,000 training steps.
\paragraph{Evaluation.} The 10\% SPIDER test split (held out from both stages) is used for all anomaly detection evaluation.
The 30\% LSMRI subset used in pre-training is disjoint from the 70\% LSMRI held-out split used to evaluate downstream report generation, ensuring no data leakage between the anomaly detector and the report generation assessment.

\paragraph{Data Augmentations}
The medium augmentation regime was selected after a three-way ablation (low/medium/high), as it reduced the intensity-distribution mismatch between best- and worst-performing validation cases  while avoiding the variance increase and performance degradation observed under aggressive augmentation (overall DICE: low~90.6, medium~90.2, high~87.8 on LSMRI/LumbarDISC localization evaluation).

\subsubsection{Heatmaps Regression Results}
\begin{table}[h]
\centering
\caption{Heatmap accuracy by dataset and training stage (medium augmentation,
mean\,$\pm$\,SD). DICE after thresholding at 0.1; RMSE on $[0,1]$-normalised
continuous values.}
\label{tab:supp-heatmap-quant-summary}
\footnotesize
\setlength{\tabcolsep}{6pt}
\resizebox{\columnwidth}{!}{%
\begin{tabular}{@{}llcc@{}}
\toprule
\textbf{Dataset} & \textbf{Stage} & \textbf{DICE} & \textbf{RMSE} \\
\midrule
LumbarDISC + LSMRI        & Localization & $90.2\pm1.0$ & $0.6\pm0.2$ \\
SPIDER (test, pos.\ pix)  & Anomaly      & $82.3\pm2.0$ & $0.4\pm0.1$ \\
\bottomrule
\end{tabular}%
}
\end{table}

\begin{figure}[h]
    \centering
    \includegraphics[width=0.88\linewidth]{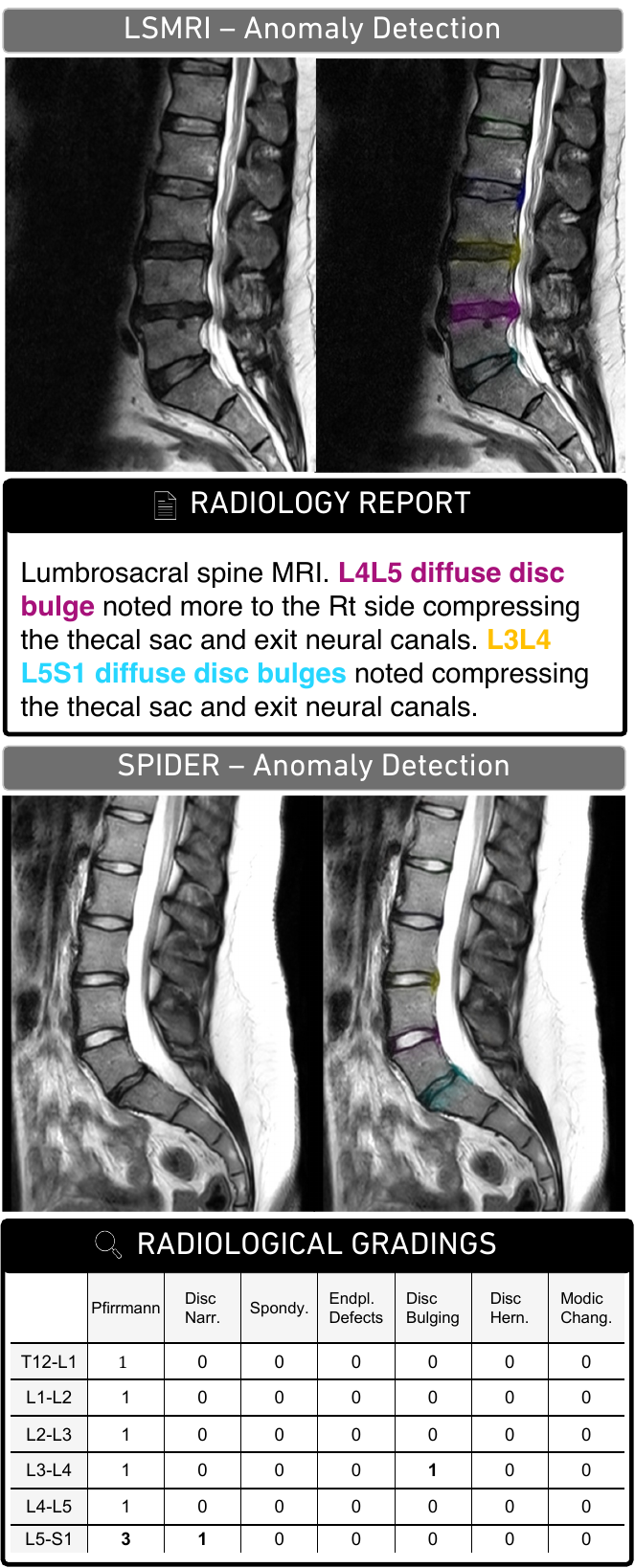}
    \caption{\textbf{Qualitative anomaly detection results on LSMRI (top) and SPIDER (bottom).} Predicted heatmaps overlaid on the central sagittal T2w slice with corresponding ground-truth annotations.
    \textit{LSMRI:} diffuse disc bulges at L3--L4 through L5--S1 (reference report); activation concentrates at lower lumbar levels.
    \textit{SPIDER:} disc bulging at L3--L4 and disc narrowing (Pfirrmann~3)  at L5--S1 (grading table); heatmap activates selectively at those levels, sparing upper IVDs.}

    \label{fig:supp-anomaly-results}
\end{figure}

Table~\ref{tab:supp-heatmap-quant-summary} summarizes the numerical heatmap results in datasets and training stages.
The drop in DICE between Phase 1 and Phase 2 (90.2 $\to$ 82.3 overall)
reflects the transition from broad disc-region targets to fine-grained
pathology-specific subregions restricting evaluation to positive
ground-truth pixels.
Performance is lowest at upper levels (T12--L1, DICE $51.8 \pm 9.2$), consistent with a lower prevalence of pathology and a lower density of supervision at those levels in the SPIDER data set. Figure~\ref{fig:supp-localization-examples} show qualitative localization heatmaps and   Figure~\ref{fig:supp-anomaly-results} shows heatmap overlay examples alongside the corresponding ground-truth annotations (clinical report for LSMRI; radiological grading for SPIDER), confirming anatomically consistent localization across IVD levels.

\begin{figure*}[tbh]
    \centering
    \includegraphics[width=0.98\linewidth]{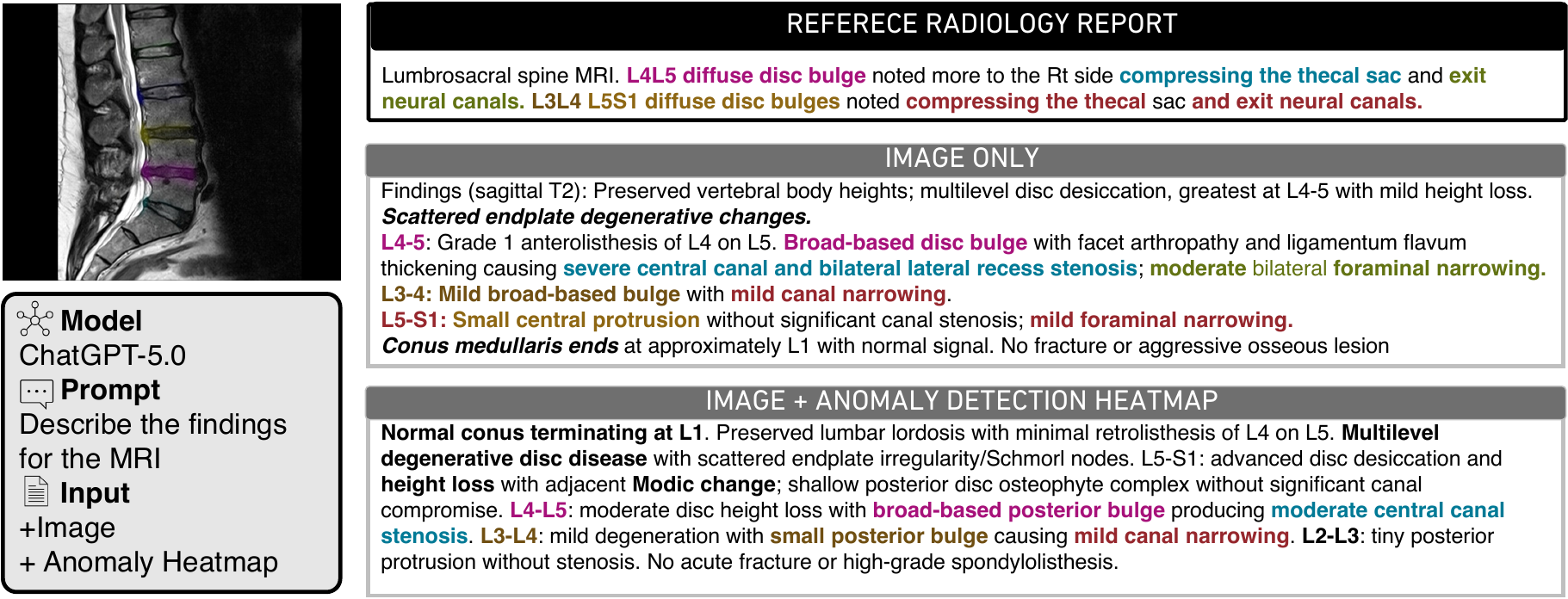}
    \caption{\textbf{Qualitative report generation results with and without anomaly detection (AD) heatmap input (ChatGPT-5.0, LSMRI).}
    The same sagittal T2w slice is shown with (\textit{img,+,AD}) and without (\textit{img only}) the predicted heatmap, alongside the clinical reference. With the heatmap, the model correctly identifies Grade~1 anterolisthesis at L4--L5, a broad-based disc bulge with severe central canal stenosis, and additional findings at L3--L4 and L5--S1, closely matching the reference. Without it, the report is clinically plausible but omits the anterolisthesis and underspecifies the severity and distribution of stenosis}
    \label{fig:supp-anomaly-guided-reports}
\end{figure*}
\subsection{Anomaly-Guided Report Generation Results}

This section provides extended evaluation of
the anomaly detection~(AD) module and its integration into report generation.
The AD module is trained with weak supervision from structured per-IVD grading
labels~(SPIDER), producing six-channel spatial heatmaps that encode
anatomically plausible pathological regions without dense annotation.
These heatmaps serve dual purposes: as standalone interpretability outputs for
direct clinical inspection, and as auxiliary visual input to ground VLM report
generation spatially.

Three input conditions are evaluated: the central sagittal slice alone
(\textit{img}), the slice augmented with the predicted heatmap overlay
(\textit{img\,+\,AD}), and an \textit{AD-guided} condition in which a
structured finding summary extracted from the heatmap is passed as textual
context — introduced for models lacking native multi-image conditioning
(MedGemma) to preserve the grounding signal within a single-image interface.
Table~\ref{tab:supp-anomaly-report-generation} reports the impact of heatmap
integration on zero-shot report generation across all evaluated models.
ChatGPT-5.0 shows consistent improvements across all metrics; MedGemma
benefits more reliably from \textit{AD-guided} than from direct overlay; and
VILA-M3 shows mixed but generally positive trends at larger model sizes.
Figure~\ref{fig:supp-anomaly-guided-reports} provides a qualitative example illustrating
how heatmap overlay promotes more anatomically specific, level-grounded reports
compared to the image-only baseline.

\begin{table}[htb]
\centering
\caption{Impact of anomaly detection (AD) heatmap overlay on zero-shot
report generation on the held-out 70\% LSMRI test split 
Models are evaluated under three input conditions: sagittal slice alone
(\textit{img}), slice augmented with the predicted disc-level heatmap
overlay (\textit{img\,+\,AD}), and AD-derived structured finding summary
passed as textual context (\textit{AD-guided}), the latter introduced for
models without native multi-image conditioning.
Performance is reported using BERTScore~F1, BLEU, ROUGE-L, and METEOR
( higher is better on 0-100 normalized sxale ).}

\label{tab:supp-anomaly-report-generation}
\resizebox{\columnwidth}{!}{%
\setlength{\tabcolsep}{6pt}
\renewcommand{\arraystretch}{1.15}
\begin{tabular}{llc cccc}
\toprule
Model & Size & Input & BERTScore & METEOR & BLEU & ROUGE-L \\
\midrule
ChatGPT & 5.0
  & img        & $91.14$ & $21.58$ & $5.45$ & $11.84$ \\
  &
  & img\,+\,AD & $\mathbf{91.60}$ & $\mathbf{23.89}$ & $\mathbf{13.41}$ & $\mathbf{23.89}$ \\

\midrule
\multirow{6}{*}{VILA-M3}
  & \multirow{2}{*}{3B}
       & img        & $\mathbf{88.89}$ & $\mathbf{5.09}$ & $0.39$ & $\mathbf{4.72}$ \\
  &    & img\,+\,AD & $88.43$ & $2.91$ & $\mathbf{0.71}$ & $4.48$ \\
\cline{2-7}
  & \multirow{2}{*}{8B}
       & img        & $\mathbf{90.78}$ & $2.79$ & $0.47$ & $4.61$ \\
  &    & img\,+\,AD & $90.05$ & $\mathbf{6.49}$ & $\mathbf{0.67}$ & $\mathbf{6.92}$ \\
\cline{2-7}
  & \multirow{2}{*}{13B}
       & img        & $87.25$ & $2.35$ & $0.19$ & $3.84$ \\
  &    & img\,+\,AD & $\mathbf{88.39}$ & $\mathbf{5.41}$ & $\mathbf{0.23}$ & $\mathbf{4.09}$ \\

\midrule
\multirow{5}{*}{MedGemma}
  & \multirow{3}{*}{4B}
       & img        & $91.09$ & $14.76$ & $5.03$ & $15.12$ \\
  &    & img\,+\,AD & $90.50$ & $9.75$  & $2.90$ & $12.44$ \\
  &    & AD-guided  & $\mathbf{91.69}$ & $\mathbf{18.28}$ & $\mathbf{12.50}$ & $\mathbf{17.27}$ \\
\cline{2-7}
  & \multirow{3}{*}{27B}
       & img        & $91.58$ & $\mathbf{22.59}$ & $\mathbf{11.86}$ & $16.32$ \\
  &    & img\,+\,AD & $91.05$ & $16.96$ & $6.76$ & $13.37$ \\
  &    & AD-guided  & $\mathbf{91.66}$ & $20.11$ & $10.96$ & $\mathbf{16.41}$ \\
\bottomrule
\end{tabular}%
}
\end{table}

\end{document}